\documentclass[runningheads]{llncs}
\usepackage{graphicx}

\usepackage{comment}
\usepackage{amsmath,amssymb} 
\usepackage{color}

\usepackage{graphicx}
\graphicspath{{figures/}}
\usepackage{subfig}

\usepackage{multirow}

\def\pt{\phantom{0}}
\definecolor{grassgreen}{rgb}{0.1,0.8,0.2}
\definecolor{lightgray}{rgb}{0.6, 0.6, 0.6}

\newcommand{\squishlist}{
 \begin{list}{$\bullet$}
  { \setlength{\itemsep}{0pt}
     \setlength{\parsep}{1pt}
     \setlength{\topsep}{1pt}
     \setlength{\partopsep}{0pt}
     \setlength{\leftmargin}{0.8em}
     \setlength{\labelwidth}{1em}
     \setlength{\labelsep}{0.5em} } }

\newcommand{\squishend}{
  \end{list}  }

\usepackage[accsupp]{axessibility}  

\begin{document}
\pagestyle{headings}
\mainmatter

\title{Synergistic Self-supervised and Quantization Learning}

\titlerunning{Synergistic Self-supervised and Quantization Learning}
%
\author{Yun-Hao Cao\inst{1} \and
Peiqin Sun\inst{2}\thanks{Corresponding author.} \and
Yechang Huang\inst{2} \and
Jianxin Wu\inst{1} \and
Shuchang Zhou\inst{2}}
\authorrunning{Y.-H. Cao et al.}
%
\institute{State Key Laboratory for Novel Software Technology, Nanjing University, China \and
MEGVII Technology \\
\email{\{caoyunhao1997, wujx2001\}@gmail.com, \{sunpeiqin, huangyechang, zsc\}@megvii.com}} 

\maketitle

\begin{abstract}
With the success of self-supervised learning (SSL), it has become a mainstream paradigm to fine-tune from self-supervised pretrained models to boost the performance on downstream tasks. However, we find that current SSL models suffer severe accuracy drops when performing low-bit quantization, prohibiting their deployment in resource-constrained applications. In this paper, we propose a method called synergistic self-supervised and quantization learning (SSQL) to pretrain quantization-friendly self-supervised models facilitating downstream deployment. SSQL contrasts the features of the quantized and full precision models in a self-supervised fashion, where the bit-width for the quantized model is randomly selected in each step. SSQL not only significantly improves the accuracy when quantized to lower bit-widths, but also boosts the accuracy of full precision models in most cases. By only training once, SSQL can then benefit various downstream tasks at different bit-widths simultaneously. Moreover, the bit-width flexibility is achieved without additional storage overhead, requiring only one copy of weights during training and inference. We theoretically analyze the optimization process of SSQL, and conduct exhaustive experiments on various benchmarks to further demonstrate the effectiveness of our method. Our code is available at \url{https://github.com/megvii-research/SSQL-ECCV2022}.
\end{abstract}

\section{Introduction}

Deep supervised learning has achieved great success in the last decade. However, traditional supervised learning approaches rely heavily on a large set of annotated training data. Self-supervised learning (SSL) has gained popularity because of its ability to avoid the cost of annotating large-scale datasets as well as the ability to obtain task-agnostic representations~\cite{sslsurvey:tang:arxiv2020}. After the emergence of the contrastive learning (CL) paradigm~\cite{simclr:hinton:ICML20,moco:kaiming:CVPR20}, SSL has clearly gained momentum and several recent works~\cite{mocov2:xinlei:arxiv2020,byol:grill:NIPS20,simsiam:kaiming:cvpr2021} have achieved comparable or even better accuracy than the supervised pretraining when transferring to downstream tasks. A standard pipeline for SSL is to learn representations (i.e., pretrained backbone networks) on unlabeled datasets and then transfer to various downstream tasks (e.g., image classification~\cite{resnet:he:CVPR16} and object detection~\cite{mask-rcnn:he:ICCV17}) by fine-tuning. 

\begin{figure*}[t]
    \centering
    \subfloat[ImageNet]{
        \label{fig:motivation-imagenet}
        \includegraphics[width=0.31\linewidth]{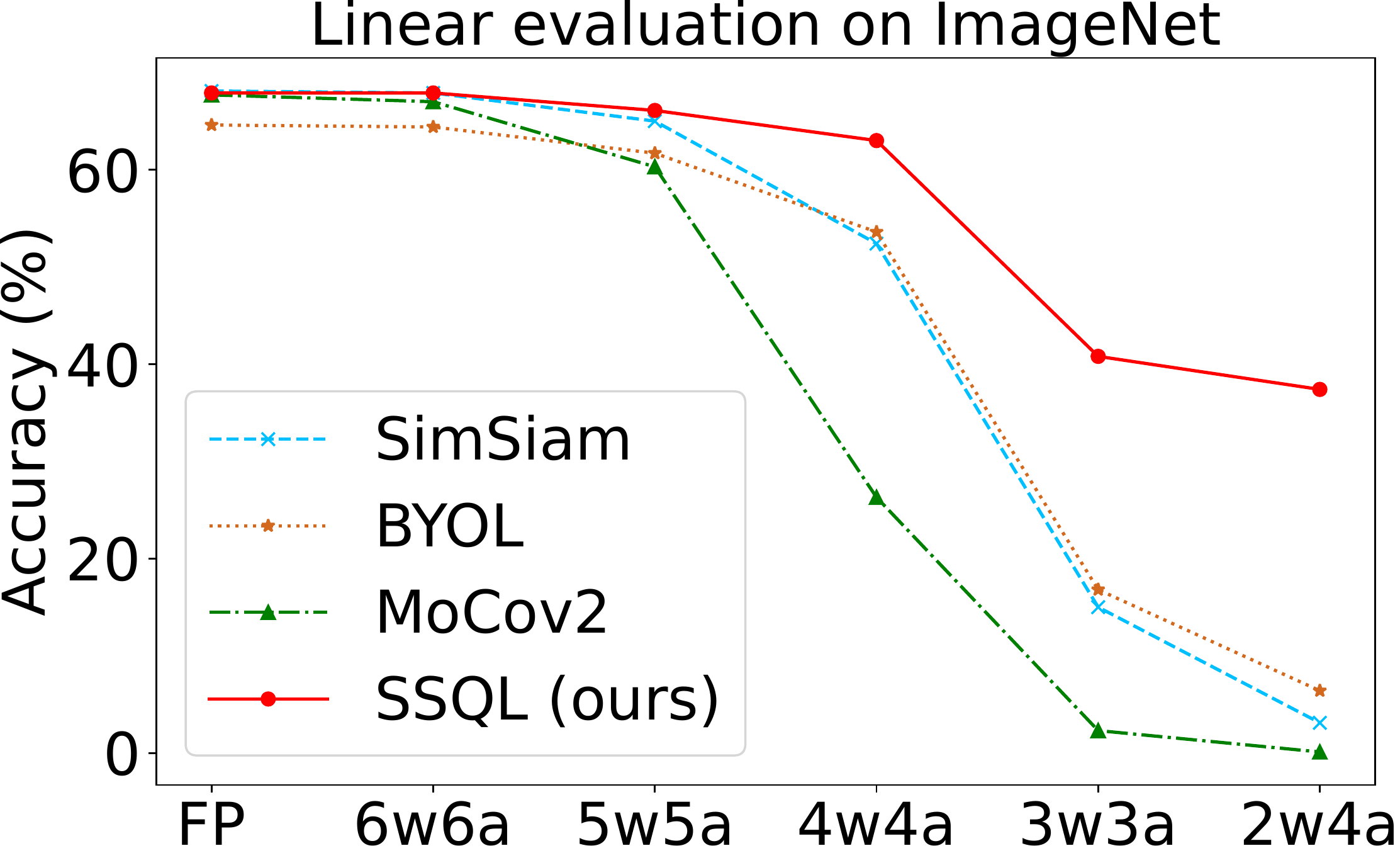}
	}
    \subfloat[Flowers]{
        \label{fig:motivation-flowers}
        \includegraphics[width=0.31\linewidth]{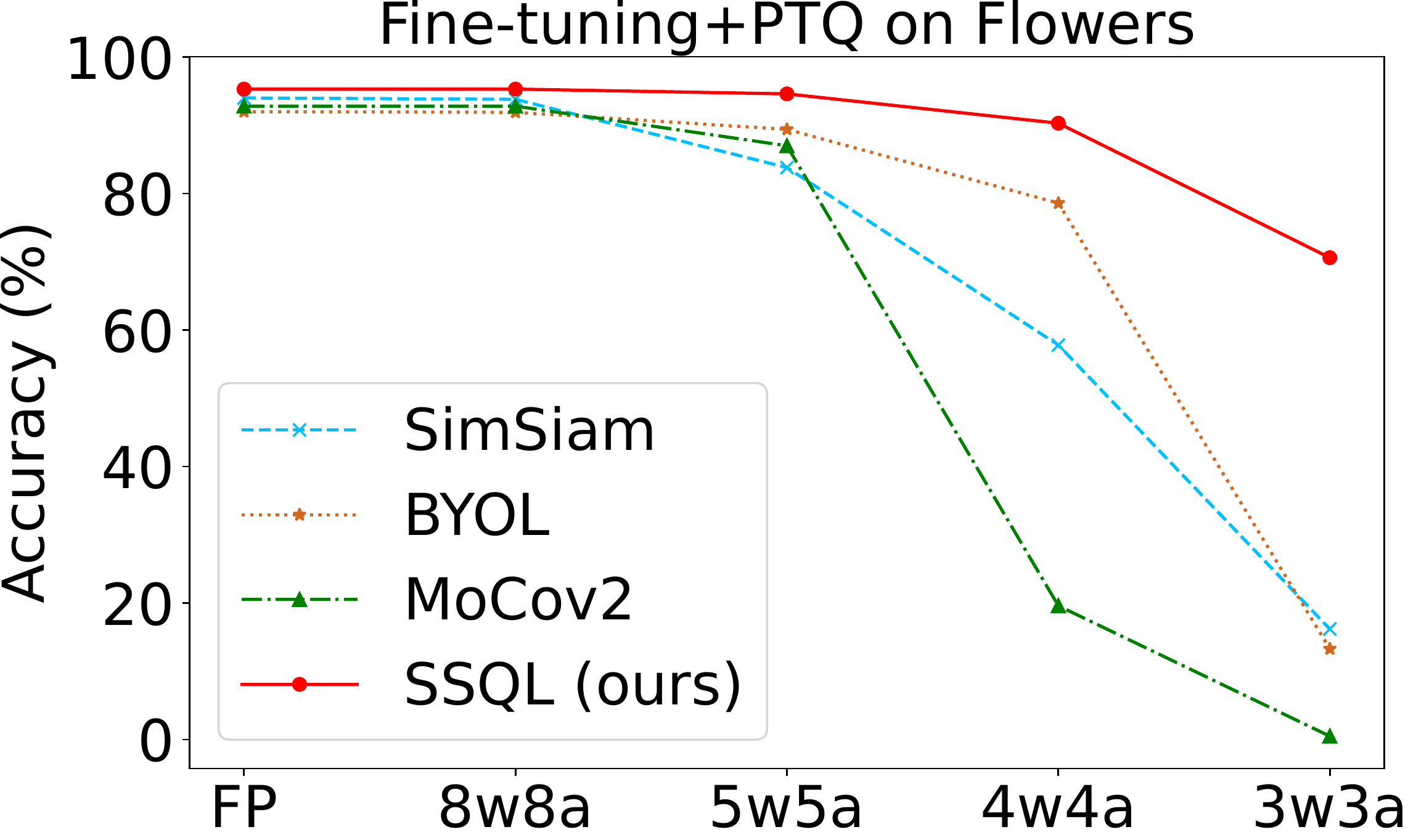}
	}
	\subfloat[COCO2017]{
        \label{fig:motivation-coco2017}
        \includegraphics[width=0.31\linewidth]{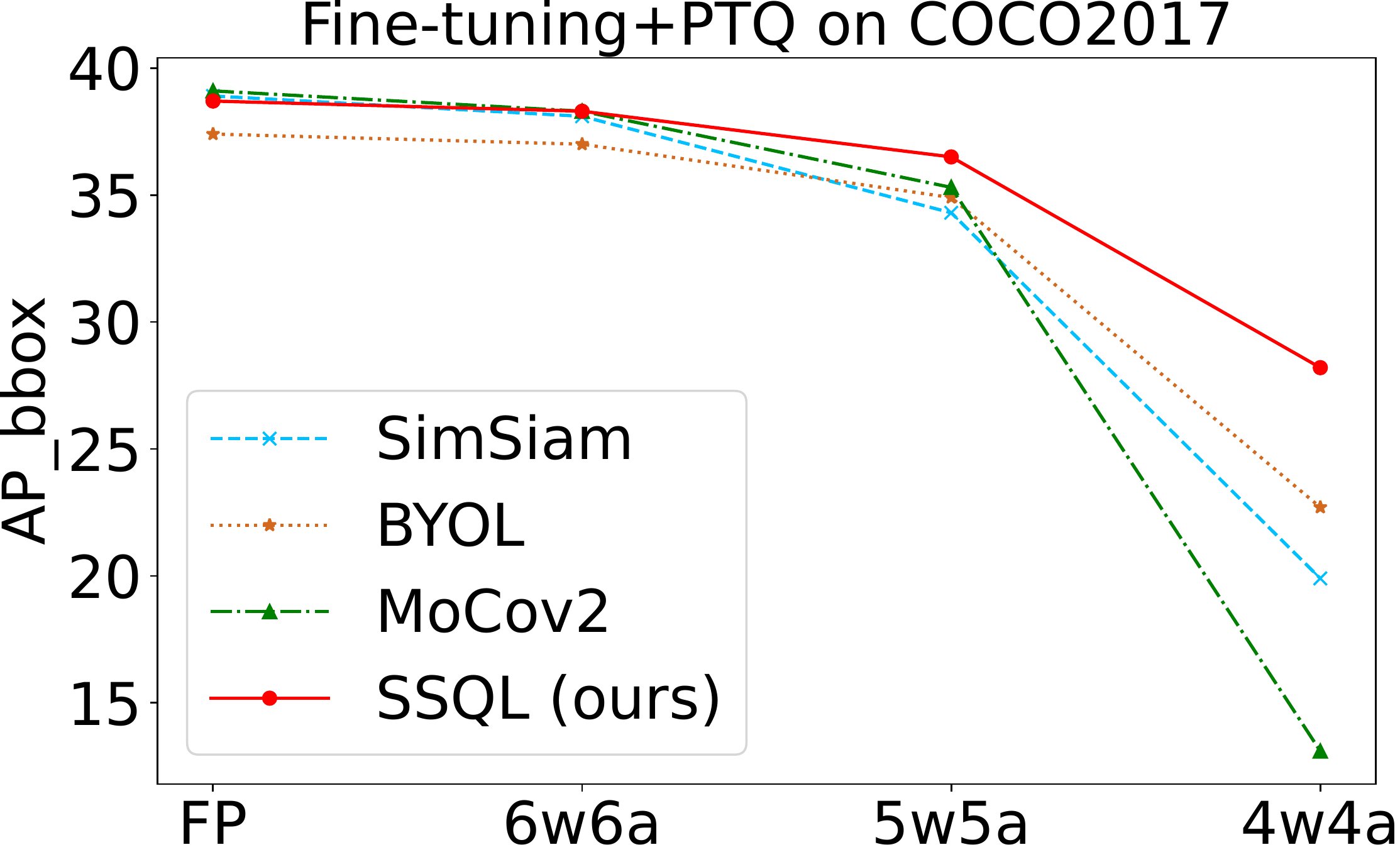}
	}
    \caption{ImageNet linear evaluation and transfer results using ImageNet pretrained models. Directly applying current self-supervised contrastive methods does not work well for low-bit quantization when transferring, while our method (SSQL) leads to a dramatic performance boost. See Section~\ref{sec:exp-imagenet} for details. `2w4a' means the weights are quantized to 2 bits and activations to 4 bits, etc.}
    \label{fig:motivation}
\end{figure*}

With the fast development of self-supervised learning, an increasing proportion of the models that need to be deployed in downstream tasks are fine-tuned from SSL pretrained models. When we want to deploy them on some resource-constrained devices, it is essential to reduce the memory consumption and latency of the neural network. To facilitate deployment, several model compression techniques have been proposed, including lightweight architecture design~\cite{mobilenetv2:sabdker:CVPR18,shufflenet:Zhang:CVPR18}, knowledge distillation~\cite{distillation:hinton:arxiv2015}, network pruning~\cite{deepcompression:han:ICLR16,thinet:luo:ICCV17}, and quantization~\cite{dorefa:zhou:arxiv2016,lsq:ICLR20}. Among them, quantization is one of the most effective methods and is directly supported by most current hardware. But severe accuracy degradation is often encountered during quantization, especially in the case of low bit-widths. As shown in Fig.~\ref{fig:motivation}, although current state-of-the-art self-supervised learning methods achieve impressive performance with full precision (FP) models, they all incur severe drop in accuracy when bit-width goes below 5. Inspired by SSL that can learn a good representation shared by various downstream tasks, we are thus motivated to ask a question: ``Can we learn a quantization-friendly representation such that the pretrained model can be quantized more easily to facilitate deployment when transferring to different downstream tasks?''.

We propose Synergistic Self-supervised and Quantization Learning (SSQL) by contrasting features of the quantized and full precision models as our solution: \emph{SSL and quantization become synergistic---they help each other}. On one hand, the contrastive loss encourages similarity of the quantized and FP models. On the other hand, quantization improves SSL by encouraging feature consistency under differently augmented weights/activations. Our contributions are:
\squishlist
    \item To the best of our knowledge, we are the first to propose quantization-friendly training for SSL. We design an effective method called SSQL, which not only greatly improves the performance when quantized to low bit-widths, but also boosts the performance of full precision models in most cases. 
    \item With SSQL, models only need to be trained \textit{once} and can then be customized for a variety of downstream tasks at different bit-widths, allowing flexible speed-accuracy trade-off for real-world deployment. The bit-width flexibility is achieved without additional storage overhead, as only \textit{one copy of weight} needs to be kept, both in the training and inference stage.
    \item SSQL is versatile. First, SSQL can be combined with existing negative-based/free CL methods. Second, the pretrained models of SSQL are compatible with existing quantization methods to further boost the performance when quantizing.
    \item We provide theoretical analysis about the synergy between SSL and quantization in SSQL. Exhaustive experimental results further show that our SSQL achieves better performance on various benchmarks at all bit-widths. 
\squishend

\section{Related Works}

\noindent\textbf{Network Quantization.} Quantization is a method that converts the weights and activations in networks from full precision (i.e., 32-bit floating-point) to fixed-point integers. According to whether or not quantization is introduced into the training process, network quantization can be divided into two categories: Quantization-Aware Training (QAT) and Post-Training Quantization (PTQ). QAT methods~\cite{dorefa:zhou:arxiv2016,pact:choi:ICLR2018,lsq:ICLR20} introduce a simulated quantization operation in the training stage. While it generally closes the gap to full precision accuracy compared to PTQ for low-bit quantization, it requires more effort in training and potentially hyperparameter tuning. In contrast, PTQ methods~\cite{adaquant:itay:ICML2021,adaround:nagel:ICML2020,brecq:ICLR21} take a trained full precision network and quantize it with little or no data~\cite{DFQ:ICCV19}, which requires minimal hyperparameter tuning and no end-to-end training. In this work, we introduce quantization into self-supervised learning to get a quantization-friendly pretrained model. Our pretrained model is compatible with existing QAT and PTQ methods when transferred to downstream tasks and hence can be combined to further improve performance. We demonstrate the quantization-friendly properties of this pretrained model through exhaustive experiments.

Adabits~\cite{adabits:CVPR20} enables adaptive bit-widths of weights and activations, but is a supervised learning method. Our pretrained model can adapt to different bit-widths, thus our work is also a method that only trains once for all bits, but in an unsupervised manner. More importantly, Adabits focuses on the current task while we investigate the transfer ability of our models and also evaluate the quantization property on downstream tasks. OQAT~\cite{oqat:ICCV21} explores extremely low-bit architecture search by combining network architecture search methods with quantization. There are also works that study quantization-friendly properties. GDRQ~\cite{GDRQ:CVPRW20} reshapes weights or activations into a uniform-like distribution dynamically. \cite{BinReg:ICCV21} proposes a bin regularization algorithm to improve low-bit network quantization. \cite{Quantfriendly-mobilenet:arxiv2018} proposes a quantization-friendly separable convolution for MobileNets. In contrast, we consider quantization-friendly properties from the perspective of pretraining under the self-supervised paradigm.

\noindent\textbf{Self-supervised Learning.} To avoid time-consuming and expensive data annotations and to explore better representations, many self-supervised methods were proposed to learn visual representations from large-scale unlabeled images or videos. Generative approaches learn to model or generate pixels in the input space~\cite{colorization:richard:ECCV16,superresolution:ledig:CVPR17,adversarial:jeff:ICLR17}. Pretext-based approaches mainly explore the context features of images or videos such as context similarity~\cite{jiasaw:mehdi:ECCV16,context:carl:ICCV15}, spatial structure~\cite{rotnet:spyros:ICLR18}, clustering property~\cite{deepclustering:caron:ECCV18}, temporal structure~\cite{sorting:lee:ICCV17}, etc. Unlike generative and pretext-based models, contrastive learning is a discriminative approach that aims at pulling similar samples closer and pushing diverse samples far from each other. Contrastive learning methods greatly improve the performance of representation learning, which have become the driving force of SSL in recent years~\cite{moco:kaiming:CVPR20,simclr:hinton:ICML20,memorybank:wu:CVPR18,byol:grill:NIPS20,swav:caron:NIPS20,simsiam:kaiming:cvpr2021}. SimCLR~\cite{simclr:hinton:ICML20} and MoCo~\cite{moco:kaiming:CVPR20} both employ a contrastive loss function InfoNCE~\cite{InfoNCE:arxiv2018}, which requires negative samples. A more radical step is made by BYOL~\cite{byol:grill:NIPS20}, which discards negative sampling in contrastive learning but achieves even better results in case a momentum encoder is used. Recently, \cite{simsiam:kaiming:cvpr2021} proposes a follow-up work SimSiam and reports surprising results that simple siamese networks can learn meaningful representations even without the momentum encoder. However, previous works did not consider whether the pretrained model is quantization-friendly when transferring to downstream tasks. 

SEED~\cite{seed:fang:ICLR21} uses self-supervised knowledge distillation for SSL with small models. $\text{S}^2$-BNN~\cite{s2bnn:CVPR21} investigates training self-supervised binary neural networks (BNN) by distilling knowledge from real networks. However, they all require a pretrained model as the teacher for distillation while ours does not. Moreover, \cite{s2bnn:CVPR21} is tailored for BNN while our method is adaptive to different bit-widths. More importantly, our method can improve the performance of the full precision (FP) model over the baseline counterpart by encouraging feature consistency
under differently augmented weights/activations via quantization.

\section{Method}

In this section, we introduce our approach, which we called synergistic self-supervised quantization learning (SSQL). We begin with the basic notation and a brief review of previous works, followed by our algorithm and analysis.

\begin{figure}[t]
	\centering
	\includegraphics[width=\columnwidth]{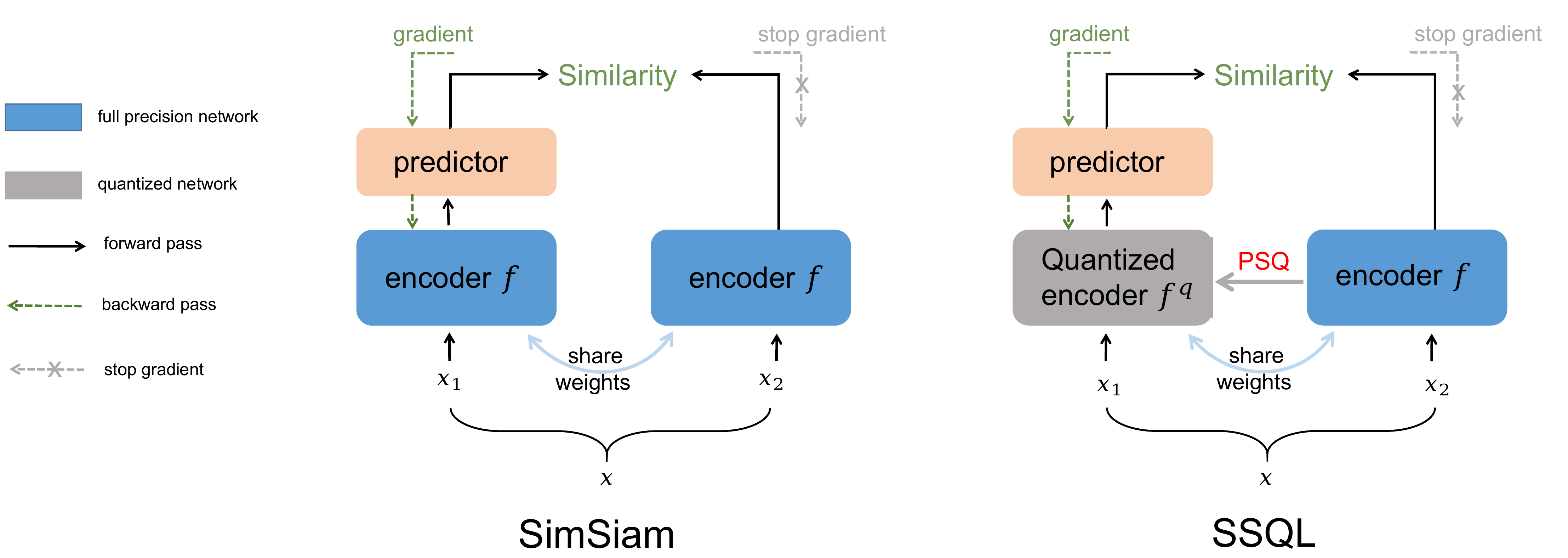}
	\caption{Illustration of our method. Left: SimSiam~\cite{simsiam:kaiming:cvpr2021}. Right: The proposed method SSQL. `PSQ' denotes post step quantization, see Sec.~\ref{sec:our-method} for details.}
	\label{fig:method}
\end{figure}

\subsection{Background and notation}

Let $\boldsymbol{x}_1$ and $\boldsymbol{x}_2$ denote two randomly augmented views from an input image $\boldsymbol{x}$. Let $f$ denote an encoder network consisting of a backbone (e.g., ResNet~\cite{resnet:he:CVPR16}) and a projection MLP head~\cite{simclr:hinton:ICML20}. By default we use SimSiam~\cite{simsiam:kaiming:cvpr2021} as the baseline counterpart to develop our algorithm, as shown in Fig.~\ref{fig:method}.

SimSiam maximizes the similarity between two augmentations of one image. A prediction MLP head~\cite{byol:grill:NIPS20}, denoted as $h$, transforms the output of one view and matches it to the other view. The output vectors for $\boldsymbol{x}_1$ is denoted as $\boldsymbol{z}_1\triangleq f(\boldsymbol{x}_1)$ and $\boldsymbol{p}_1\triangleq h(f(\boldsymbol{x}_1))$, and $\boldsymbol{z}_2$ and $\boldsymbol{p}_2$ are defined similarly.

The negative cosine similarity is defined as $D(\boldsymbol{p}, \boldsymbol{z})\triangleq -\frac{\boldsymbol{p}}{\Vert \boldsymbol{p} \Vert_2}\cdot \frac{\boldsymbol{z}}{\Vert \boldsymbol{z} \Vert_2}$ and we assume both $\boldsymbol{z}$ and $\boldsymbol{p}$ have been $l_2$-normalized for simplicity in the following. Let $SG(\cdot)$ denote the stop-gradient operation. Then, the objective to be minimized in SimSiam is then:
\begin{equation}
    \label{eq:simsiam}
    L_{\text{SimSiam}} = D(\boldsymbol{p}_1, SG(\boldsymbol{z}_2)) + D(\boldsymbol{p}_2, SG(\boldsymbol{z}_1))\,.
\end{equation}

\subsection{Our method} \label{sec:our-method}

Our motivation is to train a quantization-friendly pretrained model, hence we proposed to introduce quantization into contrastive learning. We denote $f^q$ as the quantized version of $f$, where $q$ is the assigned quantization bit-width. Correspondingly, the resulting outputs become $\boldsymbol{z}^q$ and $\boldsymbol{p}^q$. We simply adopt the commonly used uniform quantizer for both weights and activations:
\begin{equation}
\label{eq:quant}
    X_{int} = clip\left(\lfloor \frac{X}{S}+Z \rceil, 0, 2^q-1\right)\,,
\end{equation}
\begin{equation}
\label{eq:dequant}
    X_{q} = (X_{int}-Z)S \,,
\end{equation}
where $S$ (scale) and $Z$ (zero-point) are quantization parameters determined by the lower bound $l$ and the upper bound $u$ of $X$, while $X$ can be either the model weights or activations. We use minimum and maximum values for $l$ and $u$:
\begin{equation}
    l=\min(X), u=\max(X) \,,
\end{equation}
\begin{equation}
    S=\frac{u-l}{2^q-1} \,.
\end{equation}

Our solution SSQL is to let the quantized encoder $f^q$ predict the output of the full precision (FP) encoder $f$ (i.e., use FP outputs as the target):
\begin{equation}
    \label{eq:SSQL}
 L_{SSQL} = D(\boldsymbol{p}^q_1, SG(\boldsymbol{z}_2)) + D(\boldsymbol{p}^q_2, SG(\boldsymbol{z}_1))  \,.
\end{equation}

It is worth noting that we need only one copy of the model weights, which is $f$. $f^q$ can be obtained directly from $f$ using \eqref{eq:quant} and~\eqref{eq:dequant}. Further, we can add the auxiliary SimSiam loss to improve performance by combining~\eqref{eq:simsiam} and \eqref{eq:SSQL}:
\begin{equation}
\label{eq:SSQL_aux}
    L_{SSQL\text{-}aux} = L_{SimSiam} +  L_{SSQL} \,.
\end{equation}

In order to make the model quantization-friendly to different bit-widths, we \textit{randomly select} values from a set of candidate bit-widths \textit{in each step} for the assignment of $q$. In addition, we also find that this random selection operation, as a kind of augmentation, brings a performance boost. We use $2\sim8$ and $4\sim8$ bits for weight and activation, respectively, in all our experiments. Also, we quantize $f$ to get $f^q$ after each step to ensure consistency, which we name as \emph{post step quantization} (PSQ). Notice that we calculate $S$ and $Z$ during the forward pass of $f$ and hence PSQ brings negligible overhead. During the backward pass, we adopt the straight-through estimator (STE)~\cite{STE:bengoi:arxiv2013} for the quantization step. Notice that the quantized network and the floating-point network \textit{share weights}, hence when we backprop on the quantized network $W_q$ using STE, the gradients will directly operate on the floating-point network $W$. We will discuss the impact of the choice of loss functions and the candidate bit-widths set in Sec.~\ref{sec:ablation}. 

\subsection{The synergy between SSL and quantization} \label{sec:theory}

Following the notations and analyses in~\cite{simsiam:kaiming:cvpr2021}, the optimization process can be viewed as an implementation of an Expectation-Maximization (EM) like algorithm. The loss function of SSQL can be organized in the following form:
\begin{equation}
\label{eq:expectation}
    \mathcal{L}(\theta, \eta)=\mathbb{E}_{x,\mathcal{T},q}[\Vert \mathcal{F}^q_{\theta}(\mathcal{T}(x))-\eta_x \Vert^2_2]\,,
\end{equation}
where $\mathcal{F}_\theta$ is a network parameterized by $\theta$, $\mathcal{F}_\theta^q$ is obtained by quantizing $\mathcal{F}_{\theta}$, $\mathcal{T}$ is the augmentation and $x$ is an image. The expectation $\mathbb{E}[\cdot]$ is over the distribution of images, augmentations and bit-widths. $\eta_x$ is the representation of image $x$.

With the formulation of Eq.~\eqref{eq:expectation}, we consider solving
\begin{equation}
\label{eq:objective}
    \min_{\theta,\eta} \mathcal{L}(\theta,\eta) \,.
\end{equation}
The problem in~\eqref{eq:objective} can be solved by alternating between two subproblems:
\begin{equation}
    \theta^t \leftarrow \mathop{\arg\min}_{\theta} \mathcal{L}(\theta, \eta^{t-1})\,;\quad \eta^t  \leftarrow \mathop{\arg\min}_{\eta} \mathcal{L}(\theta^t, \eta)\,.
\end{equation}
Here $t$ is the index of alternation and ``$\leftarrow$'' means assigning. The optimization step for $\eta^t$ is the same as~\cite{simsiam:kaiming:cvpr2021} and we analyze the optimization step for $\theta^t$: 
\begin{equation}
\label{eq:optimize_theta}
\theta^{t+1} \leftarrow \arg\min_{\theta} \mathbb{E}_{x,\mathcal{T}, q}\Big[\Vert \mathcal{F}^q_{\theta}(\mathcal{T}(x))-\mathcal{F}_{\theta^t}(\mathcal{T}'(x)) \Vert^2_2\Big]
\end{equation}
Here $\mathcal{T}'$ implies another view and detailed derivation of~\eqref{eq:optimize_theta} is included in the appendix. Moreover, we have
\begin{align}
    &\mathbb{E}_{x,\mathcal{T},q}\Big[\Vert \mathcal{F}^q_\theta(\mathcal{T}(x))-\mathcal{F}_{\theta^t}(\mathcal{T}'(x)) \Vert^2_2\Big] \\
    =&\mathbb{E}_{x,\mathcal{T},q}\Big[\Vert \mathcal{F}^q_\theta(\mathcal{T}(x))-\mathcal{F}_\theta(\mathcal{T}(x))+\mathcal{F}_\theta(\mathcal{T}(x))-\mathcal{F}_{\theta^t}(\mathcal{T}'(x)) \Vert^2_2\Big]\\
    = &\underbrace{\mathbb{E}_{x,\mathcal{T},q}\Big[ \Vert \mathcal{F}^q_\theta(\mathcal{T}(x))-\mathcal{F}_{\theta}(\mathcal{T}(x)) \Vert^2_2 \Big]}_{\text{Q term (quantization term)}}+\underbrace{\mathbb{E}_{x,\mathcal{T},q}\Big[\Vert \mathcal{F}_\theta(\mathcal{T}(x))-\mathcal{F}_{\theta^t}(\mathcal{T'}(x)) \Vert^2_2 \Big]}_{\text{CL term (contrastive learning term)}}\\
    &+\underbrace{2\mathbb{E}_{x,\mathcal{T},q}\Big[ \big(\mathcal{F}^q_\theta(\mathcal{T}(x))-\mathcal{F}_{\theta}(\mathcal{T}(x))\big)^T\big(\mathcal{F}_\theta(\mathcal{T}(x))-\mathcal{F}_{\theta^t}(\mathcal{T'}(x))\big) \Big]}_{\text{cross term}}
\end{align}

It is reasonable to assume that the quantization error and the contrastive learning error are at most weakly correlated\footnote[1]{We also empirically verify this assumption in the appendix.}, hence we can remove the cross term and are left with two objectives in the optimization step for $\theta$. The Q term minimizes the distance between the quantized network $\mathcal{F}^q_{\theta}$ and the FP network $\mathcal{F}_{\theta}$, which naturally leads to the desired quantization-friendly property. The CL term is the original optimization term in SimSiam to learn image representations. Also notice that we take expectations over 3 terms, where the extra $q$ term can be seen as one kind of augmentation on weights/activations. It is well-known that strong image augmentations are essential in SSL~\cite{simclr:hinton:ICML20}. Hence, the quantization can potentially assist the learning of SSL, by encouraging feature consistency under differently augmented weights/activations via quantization. In conclusion, the design of our loss function makes quantization and SSL work in a synergistic fashion, i.e., they help each other.

\section{Experiments}

We introduce the implementation details in Sec.~\ref{sec:details}. We experiment on CIFAR-10 and CIFAR-100~\cite{cifar} in Sec.~\ref{sec:exp-cifar} and ImageNet~\cite{ILSVRC2012:russakovsky:IJCV15} (IN) in Sec.~\ref{sec:exp-imagenet}. Then, we evaluate the transfer performance of ImageNet pretrained models on downstream classification and object detection benchmarks in Sec.~\ref{sec:exp-imagenet}. Finally, we study the effects of different components and hyper-parameters in our algorithm in Sec.~\ref{sec:ablation}. All our experiments were conducted using PyTorch~\cite{pytorch:NIPS19}. Codes will be available.

\subsection{Implementation details} \label{sec:details}

\textbf{Datasets.} The main experiments are conducted on three benchmark datasets, i.e., CIFAR-10, CIFAR-100 [29] and ImageNet~\cite{ILSVRC2012:russakovsky:IJCV15}. The large-scale ImageNet contains 1.28M images for training from 1,000 classes. We also conduct transfer experiments on 7 recognition benchmarks (see appendix for details) as well as 2 detection benchmarks Pascal VOC 07\&12~\cite{VOC:mark:IJCV10} and COCO2017~\cite{coco:LinTY:ECCV14}.

\noindent\textbf{Backbones.} Apart from the commonly used ResNet-50~\cite{resnet:he:CVPR16} in recent SSL papers, we also adopt 2 smaller networks, i.e., ResNet-18~\cite{resnet:he:CVPR16} and ResNet-34~\cite{resnet:he:CVPR16} for our experiments. We use the same settings as \cite{simsiam:kaiming:cvpr2021} for prediction and projection MLP. Sometimes we abbreviate ResNet-18/50 to R-18/50.

\noindent\textbf{Training details.} We follow the training setup in SimSiam~\cite{simsiam:kaiming:cvpr2021} for our method. More specifically, we use SGD for pretraining, with batch size of 256 and a base lr=0.05. The learning rate has a cosine decay schedule. The weight decay is 0.0001 and the SGD momentum is 0.9. We pretrain for 400 epochs on CIFAR-10 and CIFAR-100 and 100 epochs on ImageNet unless otherwise specified. 

For ImageNet linear evaluation, we follow the same settings in \cite{simsiam:kaiming:cvpr2021}. For linear evaluation on other datasets, we train for 100 epochs with lr initialized to 30.0, which is divided by 10 at the 60th and 80th epoch. For fine-tuning, we train for 50 epochs with lr initialized to 0.001, which is divided by 10 at the 30th and 40th epoch. The weight decay is 0 for linear evaluation and 1e-4 for fine-tuning. 

\begin{figure}[t]
	\centering
	\includegraphics[width=0.95\columnwidth]{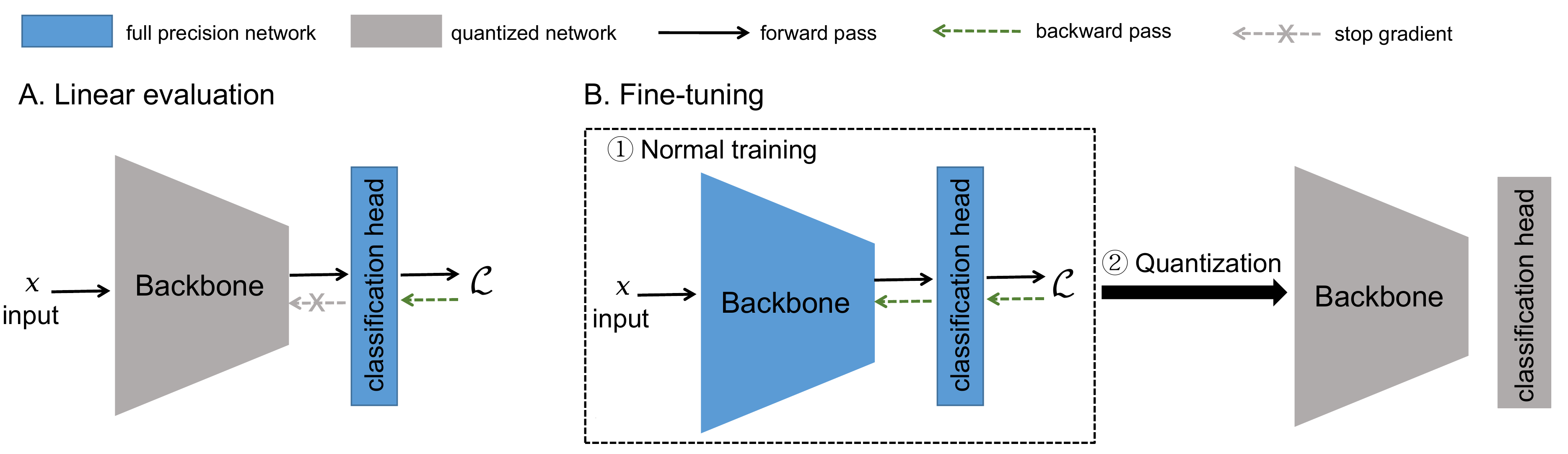}
	\caption{Illustration of the evaluation protocols adopted in our paper.}
	\label{fig:protocol}
\end{figure}

\noindent\textbf{Evaluation protocols.} Following previous works~\cite{moco:kaiming:CVPR20}, we adopt linear evaluation and fine-tuning to evaluate the pretrained representations. Moreover, we want to evaluate the performance of the representations after quantization. Hence, we make corresponding adjustments and propose a new evaluation protocol when combining quantization and SSL, as shown in Fig.~\ref{fig:protocol}. More specifically, we freeze and quantize the backbone and only update the classification head for linear evaluation (i.e., backbone weights frozen). For fine-tuning, we first train the backbone as well as the classification head as normal (i.e., backbone weights updated). Then, based on the fine-tuned FP model, we conduct either PTQ or QAT to evaluate the performance after quantization. We adopt PTQ after fine-tuning in our experiments by default. We use `$n$w$m$a' to denote that we quantize weight to $n$-bit and quantize activation to $m$-bit in this paper (e.g., 4w4a).

\begin{table}[t]
	\caption{Linear evaluation results on CIFAR-10. All pretrained for 400 epochs. SimSiam-PACT trains 7 models separately and we color it grey.}
	\label{tab:cifar10}
	\centering
	\renewcommand{\arraystretch}{0.85}
	\renewcommand{\multirowsetup}{\centering}
	\begin{tabular}{l|c|c|c|c|c|c|c|c|c}
		\hline
		\multirow{2}{*}{Backbone} & \multirow{2}{*}{Method} & \multicolumn{8}{c}{Linear evaluation accuracy (\%)} \\
		\cline{3-10}
		& & FP & 8w8a & 6w6a & 5w5a & 4w4a & 3w3a & 2w8a & 2w4a\\
		\hline
		\multirow{7}{*}{ResNet-18}& SimSiam~\cite{simsiam:kaiming:cvpr2021} & \textbf{90.7} & 90.7 & \textbf{90.6} & 90.3 & 88.9 & 66.0 & 70.1 & 63.8\\
		& BYOL~\cite{byol:grill:NIPS20} & 89.3 & 89.3 & 89.4 & 89.3 & 88.0& 75.1 & 71.9 & 63.3 \\
		& SimSiam-PACT~\cite{pact:choi:ICLR2018} & - &\color{lightgray}{89.2}&\color{lightgray}{89.2}&\color{lightgray}{89.3} & \color{lightgray}{89.2}&\color{lightgray}{88.2}&\color{lightgray}{89.3}&\color{lightgray}{88.3}\\
		& SSQL (ours) & \textbf{90.7} & \textbf{90.8} & \textbf{90.6} & \textbf{90.6} & \textbf{90.1}& \textbf{85.6} & \textbf{88.0} & \textbf{86.5} \\
		\cline{2-10}
		& SimCLR~\cite{simclr:hinton:ICML20} &\textbf{89.4} & \textbf{89.3}  & \textbf{89.2} & \textbf{88.8} & 87.1& 73.9 & 65.6 & 55.6  \\
		& MoCov2~\cite{mocov2:xinlei:arxiv2020} &88.9 & 88.8 & 88.4 & 88.2 & 86.8& 72.2 & 66.4 & 50.7  \\
		& SSQL-NCE (ours) & 89.0 &89.0 & 89.0 & \textbf{88.8}  &\textbf{87.9} & \textbf{82.9} & \textbf{87.1} & \textbf{84.9} \\
		\hline
		\multirow{6}{*}{ResNet-50}& SimSiam~\cite{simsiam:kaiming:cvpr2021} & 90.9	& 90.9	&	91.0& 90.6 &  89.5	& 	74.1 & 55.1& 57.1  \\
		& BYOL~\cite{byol:grill:NIPS20} & 90.3 & 90.3 & 90.0 & 89.7 & 87.5 & 58.5 & 82.4 & 67.8 \\
		& SSQL (ours) & \textbf{91.1} & \textbf{91.1}	& 	\textbf{91.1} & \textbf{91.1} & 	\textbf{90.0} & \textbf{77.4} &	\textbf{89.5} & \textbf{87.2}	 \\
		\cline{2-10}
		& SimCLR~\cite{simclr:hinton:ICML20} & 91.5 & 91.4 & 91.3 & 90.5 & 88.1 & 59.6 & 63.5 & 42.4\\
		& MoCov2~\cite{mocov2:xinlei:arxiv2020} & 90.2 & 90.2 & 90.2 & 89.4 & 87.9 & 72.1 & 68.8 & 49.5 \\
		& SSQL-NCE (ours) &  \textbf{92.1}  & \textbf{92.1} & \textbf{92.0} &  \textbf{91.9} & \textbf{89.8} & \textbf{74.0} & \textbf{88.6} & \textbf{84.9} \\
		\hline  
	\end{tabular}
\end{table}

\subsection{CIFAR results} \label{sec:exp-cifar}

We compare our method with popular SSL methods BYOL~\cite{byol:grill:NIPS20}, SimSiam~\cite{simsiam:kaiming:cvpr2021}, SimCLR~\cite{simclr:hinton:ICML20} and MoCov2~\cite{mocov2:xinlei:arxiv2020}. We evaluate the linear evaluation accuracy under different bit-widths after quantization, as mentioned in Sec~\ref{sec:details}. Notice that we only pretrain one full precision (FP) model and then use it for evaluation on different bit-widths. To better illustrate the effectiveness of our method, we also create one strong baseline SimSiam-PACT, by combining PACT~\cite{pact:choi:ICLR2018} and SimSiam during pretraining. Notice that it is not a fair comparison with other methods because it needs to pretrain different models for different bit-widths (i.e., need 7 pretrained models for 7 bit-widths). In other words, it is not flexible and the training overhead is unbearable for large data volumes. Experimental results on CIFAR-10 and CIFAR-100 are shown in Table~\ref{tab:cifar10} and Table~\ref{tab:cifar100}, respectively. 

As shown in Table~\ref{tab:cifar10}, take ResNet-18 as an example, our SSQL achieves comparable performance with the baseline counterpart SimSiam under linear evaluation in full precision on CIFAR-10. However, when we lower the bit-width (from 8w8a to 2w4a), our advantages over the baseline SimSiam will become more and more obvious. For instance, our SSQL achieves \textbf{19.6\%} and \textbf{22.7\%} higher accuracy than SimSiam at 3w3a and 2w4a, respectively. When comparing with SimSima-PACT, we can find that our SSQL achieves higher accuracy at 4w4a and above. However, SimSiam-PACT achieves slightly higher accuracy than our method at 3w3a and below but the gap is within 3\%. Moreover, we achieve higher accuracy than SimSiam under ResNet-50 at FP, and the advantages when reducing bit-widths are consistent. Finally, our SSQL can also be combined with InfoNCE~\cite{InfoNCE:arxiv2018} based methods, e.g., SimCLR and we name it SSQL-NCE. We can observe similar trends as above and it demonstrates that our SSQL is compatible with both negative-based and negative-free CL methods.

\begin{table}[t]
	\caption{Linear evaluation results on CIFAR-100. All pretrained for 400 epochs.}
	\label{tab:cifar100}
	\centering
	\renewcommand{\arraystretch}{0.85}
	\renewcommand{\multirowsetup}{\centering}
	\begin{tabular}{l|c|c|c|c|c|c|c|c|c}
		\hline
		\multirow{2}{*}{Backbone} & \multirow{2}{*}{Method} & \multicolumn{8}{c}{Linear evaluation accuracy (\%)} \\
		\cline{3-10}
		& & FP & 8w8a & 6w6a & 5w5a & 4w4a & 3w3a & 2w8a & 2w4a\\
		\hline
		\multirow{5}{*}{ResNet-18}& SimSiam~\cite{simsiam:kaiming:cvpr2021} &65.5& 65.5	&65.4 & 64.6 & 62.6&	41.6 & 40.1& 36.9  \\
		& BYOL~\cite{byol:grill:NIPS20} &62.6 &62.6&62.5&62.0&60.6&47.9&44.1&38.8\\
		& SimCLR~\cite{simclr:hinton:ICML20} & 59.2 &59.2 & 59.0 & 57.9 & 54.4 & 34.1 &38.4  &28.8   \\
		& MoCov2~\cite{mocov2:xinlei:arxiv2020} & 62.5 & 62.5 & 62.1 & 61.5 & 59.5 & 43.5 & 40.1 & 30.8 \\
		& SSQL (ours) &\textbf{66.9}& \textbf{66.8}&\textbf{66.9}& \textbf{65.8} & \textbf{65.0}&\textbf{57.4}& \textbf{53.9}& \textbf{50.6}   \\
		\hline
		\multirow{5}{*}{ResNet-50}& SimSiam~\cite{simsiam:kaiming:cvpr2021} & 64.3	& 64.2&64.1& 62.9 & 61.3	& 44.9 & 32.9& 32.6	   \\
		& BYOL~\cite{byol:grill:NIPS20} & 66.7 &66.5 &65.0&59.6&47.2&14.5&55.3&27.2 \\
		& SimCLR~\cite{simclr:hinton:ICML20} & 66.2 & 66.1 & 65.9 & 64.8& 60.1 & 40.2 & 43.8 & 24.7   \\
		& MoCov2~\cite{mocov2:xinlei:arxiv2020} & 66.5 &66.5&66.3 & 65.4 & 61.9 &  44.2&41.1 &28.5 \\
		& SSQL (ours) &  \textbf{68.0} & \textbf{67.9} & \textbf{67.8} & \textbf{67.8} & \textbf{67.8}	& \textbf{59.9} & \textbf{62.9} & \textbf{61.5} \\
		\hline 
	\end{tabular}
\end{table}

As shown in Table~\ref{tab:cifar100}, our SSQL achieves the highest accuracy on CIFAR-100 in all cases. For instance, when comparing the first column (FP), our SSQL is significantly better than baseline counterpart SimSiam: up to \textbf{+1.4\%} and \textbf{+3.7\%} accuracy for ResNet-18 and ResNet-50, respectively. Our advantages become bigger when we further lower the bit-widths: up to \textbf{+6.5\%}, \textbf{+15\%} and \textbf{+28.9\%} accuracy at 4w4a, 3w3a and 2w4a, respectively, for ResNet-50. 

\begin{figure}[t]
	\centering
	\includegraphics[width=\columnwidth]{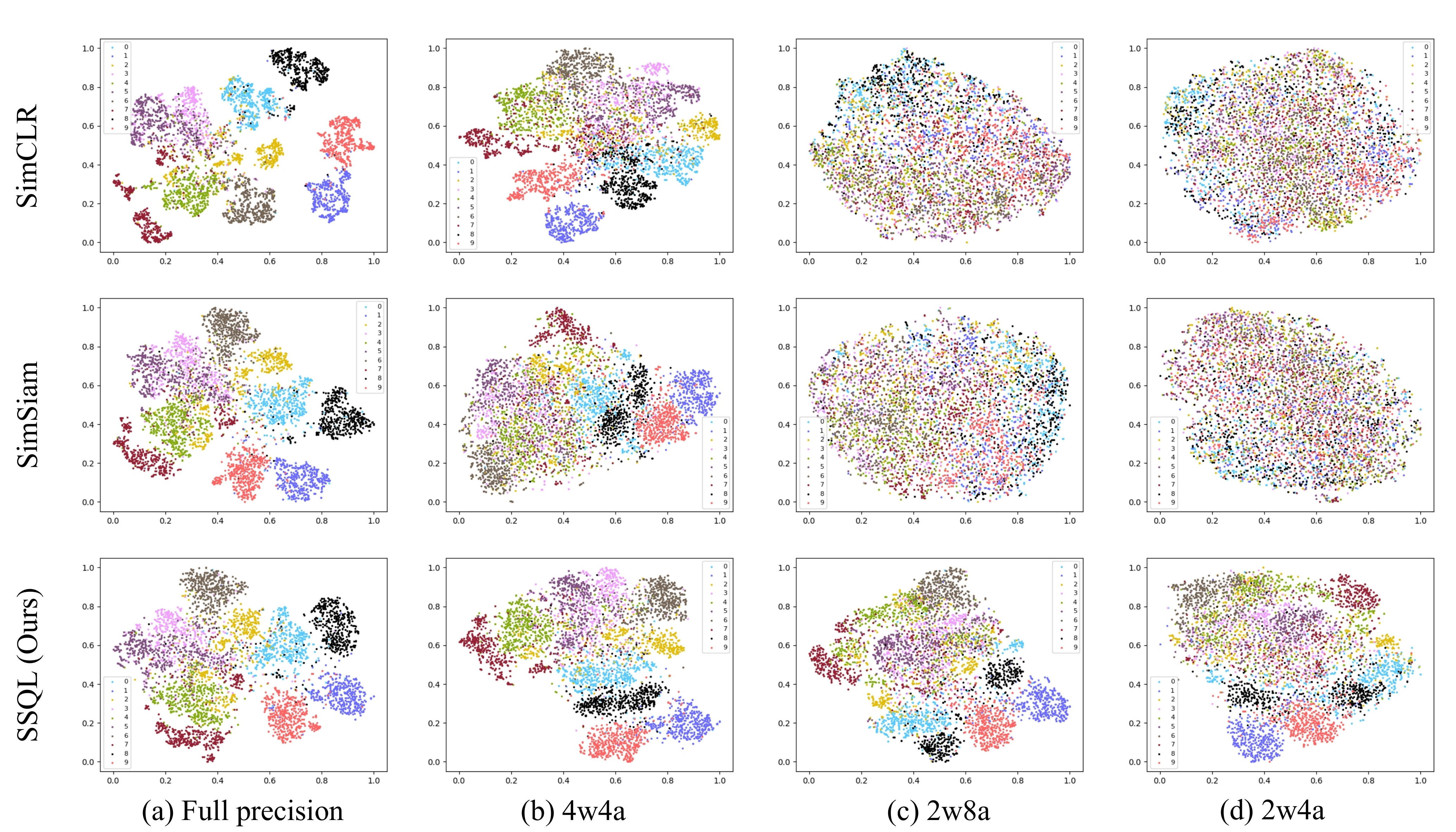}
	\caption{t-SNE~\cite{tsNE} visualization of CIFAR-10 using ResNet-18. The column (a) shows the results using FP backbone. The column (b), (c) and (d) shows the results at 4w4a, 2w8a and 2w4a, respectively. This figure is best viewed in color.}
	\label{fig:tSNE}
\end{figure}

To demonstrate the effectiveness of the proposed method in a more intuitive way, we visualize the feature spaces learned by different methods in Fig.~\ref{fig:tSNE}. First, three models are trained on the CIFAR-10 dataset by using SimCLR, SimSiam and SSQL, respectively. After that, 5,000 samples in CIFAR-10 are represented accordingly and then are reduced to a two-dimensional space by t-SNE~\cite{tsNE}. As seen, the samples are more separable in the feature space learned by SSQL than both SimCLR and SimSiam (especially at 2w8a and 2w4a), showing that SSQL can learn better feature representations after quantization.

\begin{table}[t]
	\caption{Linear evaluation results on ImageNet. All pretrained for 100 epochs, except for MoCov2. $^\dagger$ denotes that we use the official MoCov2 200ep checkpoint. SimSiam-PACT trains 5 models separately and we color it grey.}
	\label{tab:imagenet_linear}
 	\centering
 	\renewcommand{\arraystretch}{0.85}
  	\renewcommand{\multirowsetup}{\centering}
	\begin{tabular}{l|c|c|c|c|c|c|c}
		\hline
		\multirow{2}{*}{Backbone} & \multirow{2}{*}{Method} & \multicolumn{6}{c}{Linear evaluation accuracy (\%)} \\
		\cline{3-8}
		& & FP & 8w8a & 5w5a & 4w4a & 3w3a & 2w4a\\
		\hline
		\multirow{4}{*}{ResNet-18}& SimSiam~\cite{simsiam:kaiming:cvpr2021} &55.0 & 54.7 & 53.9 & 36.7 & \phantom{0}6.3 & \phantom{0}1.5   \\
		& BYOL~\cite{byol:grill:NIPS20} & 54.1 &54.0&51.9&42.4&13.6&\phantom{0}3.6 \\
		& SimSiam-PACT~\cite{pact:choi:ICLR2018} & - &\textcolor{lightgray}{52.8} & \textcolor{lightgray}{52.8} &  \textcolor{lightgray}{52.3} & \textcolor{lightgray}{51.0} & \textcolor{lightgray}{51.6} \\
		& SSQL (ours) & \textbf{57.6} & \textbf{57.6}& \textbf{56.7} & \textbf{52.8} & \textbf{41.0} & \textbf{43.1}   \\
		\hline
		\multirow{4}{*}{ResNet-50}& SimSiam~\cite{simsiam:kaiming:cvpr2021} & \textbf{68.1} & \textbf{67.9} & 65.0 & 52.4 &15.0 & \phantom{0}3.1  \\
		& BYOL~\cite{byol:grill:NIPS20} &64.6 &64.4&61.7&53.6&16.8 & \phantom{0}6.4\\
		& MoCov2$^\dagger$~\cite{mocov2:xinlei:arxiv2020} & 67.7 & 67.0& 60.3& 26.3 &  \phantom{0}2.3 & \phantom{0}0.1\\
		& SSQL (ours) &  67.9 & \textbf{67.9} & \textbf{66.1} & \textbf{63.0} & \textbf{40.8} &  \textbf{37.4} \\
		\hline 
	\end{tabular}
\end{table}

\subsection{ImageNet and transfer learning results} \label{sec:exp-imagenet}

In this section, we do unsupervised pretraining on the large-scale ImageNet training set~\cite{ILSVRC2012:russakovsky:IJCV15} without using labels. The linear evaluation results on ImageNet are shown in Table~\ref{tab:imagenet_linear}. Also, we evaluate the transfer ability of the learned representations on ImageNet later. We train SSQL, SimSiam and BYOL for 100 epochs on ImageNet and directly use the official checkpoint for MoCov2.

As shown in Table~\ref{tab:imagenet_linear}, when comparing the first column (FP), our SSQL achieves higher accuracy than the baseline counterpart SimSiam (57.6 v.s. 55.0) under ResNet-18. When comparing the fourth column (4w4a), our SSQL achieves 16.1\% and 10.6\% gains for ResNet-18 and ResNet-50, respectively. In short, our SSQL achieves comparable or better accuracy at full precision and is more quantization-friendly at lower bit-widths. When compared with SimSiam-PACT, our SSQL achieves better results at 4 bits or higher, \textit{with only one copy of weights}.

\begin{figure}[t]
	\centering
	\includegraphics[width=0.95\columnwidth]{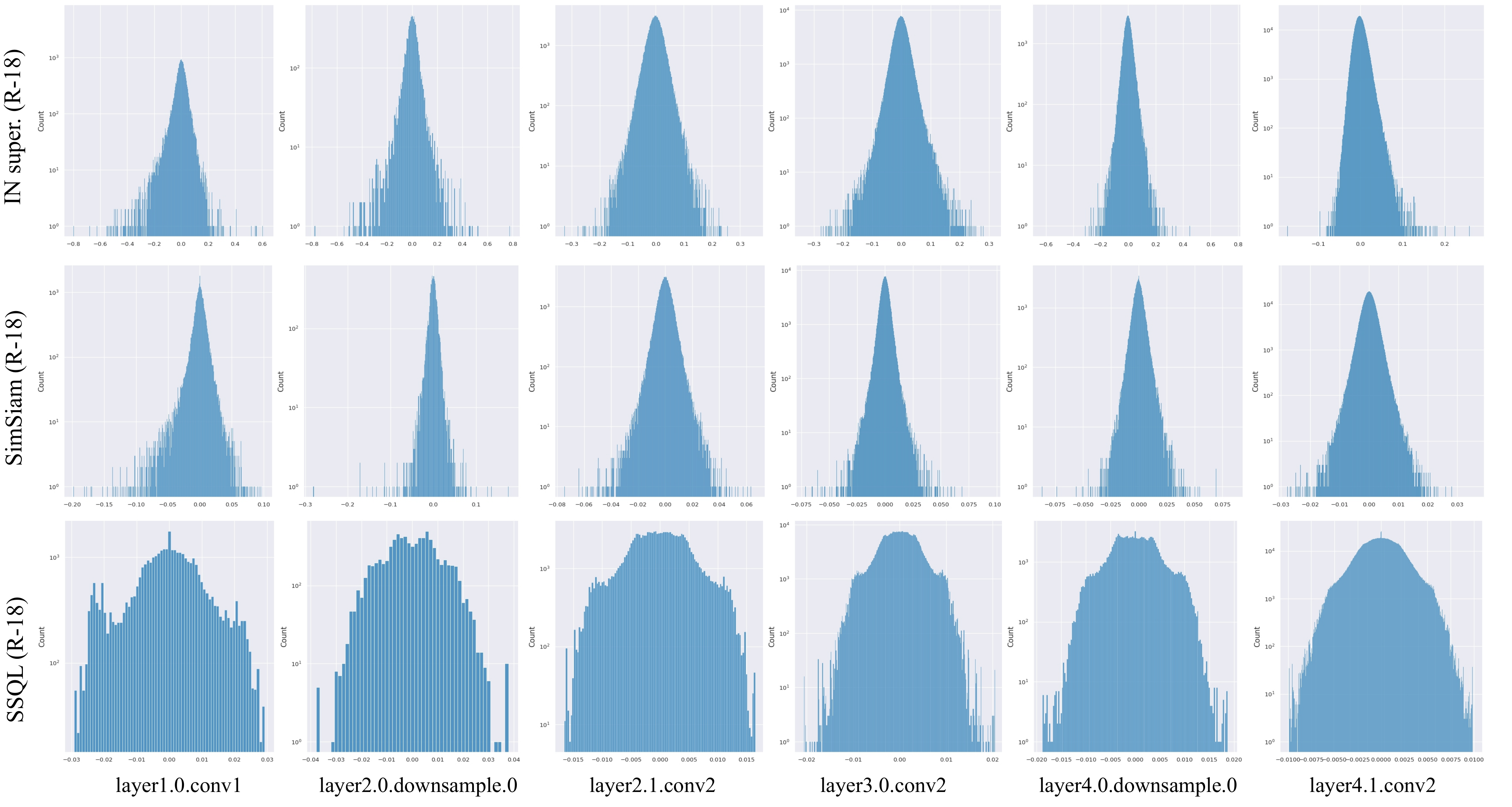}
	\caption{Visualization of weight distribution for ResNet-18. The first, second and third row are the results of ImageNet supervised, SimSiam and ours, respectively.}
	\label{fig:weight-R-18}
\end{figure}

As shown in Fig.~\ref{fig:motivation}, the ImageNet linear evaluation performance can somehow indicate the performance at downstream tasks at different bit-widths (i.e., the trend is consistent). We plot the weight distribution of different pretrained models in Fig.~\ref{fig:weight-R-18}. As seen, the weights of our model (third row) are more quantization-friendly when compared with the two baseline counterparts in terms of 3 aspects: more uniform distribution, smaller ranges, and much fewer outliers. (There is a similar phenomenon after fine-tuning on downstream tasks, too, see appendix). 

\begin{table}[t]
	\caption{Fine-tuning+PTQ results on ImageNet subsets. Here we adopt the fine-tuning settings on 1\%/10\% labeled data and report Top-5 accuracy (\%).}
	\label{tab:imagenet_finetune}
 	\centering
 	\renewcommand{\arraystretch}{0.9}
  	\renewcommand{\multirowsetup}{\centering}
	\begin{tabular}{l|c|c|c|c|c|c|c|c|c}
		\hline
		\multirow{2}{*}{Backbone} & \multirow{2}{*}{Method} & \multicolumn{4}{c|}{1\% labels} & \multicolumn{4}{c}{10\% labels}\\
		\cline{3-10}
		& & FP & 6w6a & 5w5a & 4w4a & FP & 6w6a & 5w5a & 4w4a \\
		\hline
		\multirow{3}{*}{ResNet-18}& SimSiam~\cite{simsiam:kaiming:cvpr2021} &43.7 &43.4&42.4&37.5&\textbf{76.1}&75.8&74.3&64.5  \\
		& BYOL~\cite{byol:grill:NIPS20} & 36.7&36.5&35.5&31.2& 75.5&75.1&73.9& 65.2\\
		& SSQL (ours) &\textbf{47.7} &\textbf{47.6}&\textbf{47.1}&\textbf{45.0}& \textbf{76.1}&\textbf{75.9}& \textbf{75.0} & \textbf{70.7}   \\
		\hline
		\multirow{3}{*}{ResNet-50}& SimSiam~\cite{simsiam:kaiming:cvpr2021}  & 53.2&52.8&51.5&36.4& 82.5 & 81.7 & 79.0 & 67.9   \\
		& BYOL~\cite{byol:grill:NIPS20} & 47.3&47.2&46.4&40.4&81.1&80.7&79.6&69.9\\
		& SSQL (ours)  & \textbf{55.2}&\textbf{55.0}&\textbf{54.4}&\textbf{51.8}& \textbf{83.0} & \textbf{82.7} & \textbf{81.0} & \textbf{76.7}    \\
		\hline 
	\end{tabular}
\end{table}

\noindent\textbf{Fine-tuning with partial labels.} Following previous practices, we also fine-tune the pretrained models with a randomly initialized linear classifier on ImageNet with 1\% and 10\% labeled data. As shown in Table~\ref{tab:imagenet_finetune}, our SSQL achieves the best performance in all cases. We also report the PTQ performance and we can see that our advantages become greater as the bit-width decreases. For instance, when fine-tuned using 10\% labels under R-50, our method achieves 0.5\% and 8.8\% higher accuracy than SimSiam at FP and 4w4a, respectively.

\begin{figure}[t]
    \centering
    \subfloat[4w4a]{
        \label{fig:LSQ-4w4a}
        \includegraphics[width=0.3\linewidth]{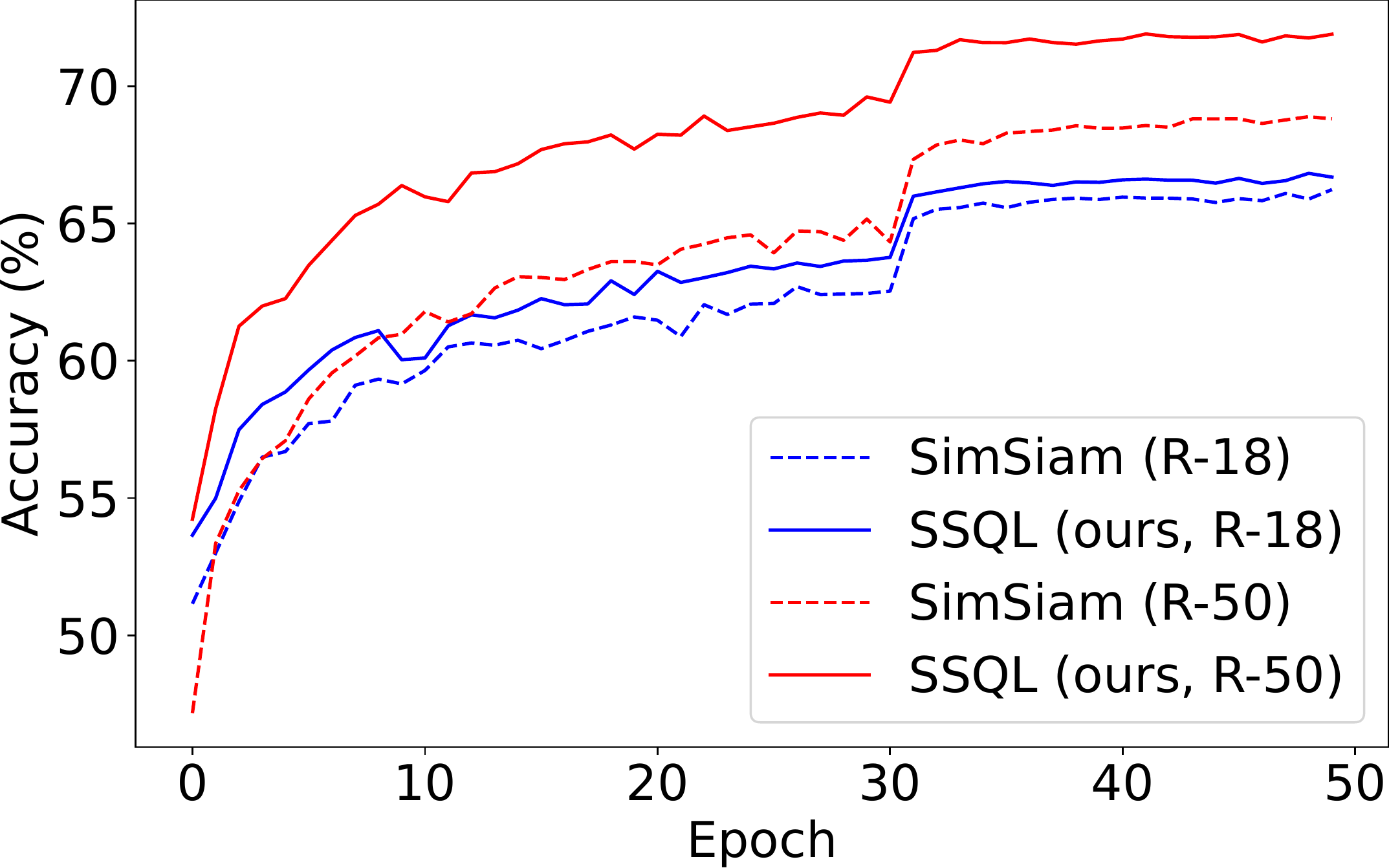}
	}
    \subfloat[3w3a]{
        \label{fig:LSQ-3w3a}
        \includegraphics[width=0.3\linewidth]{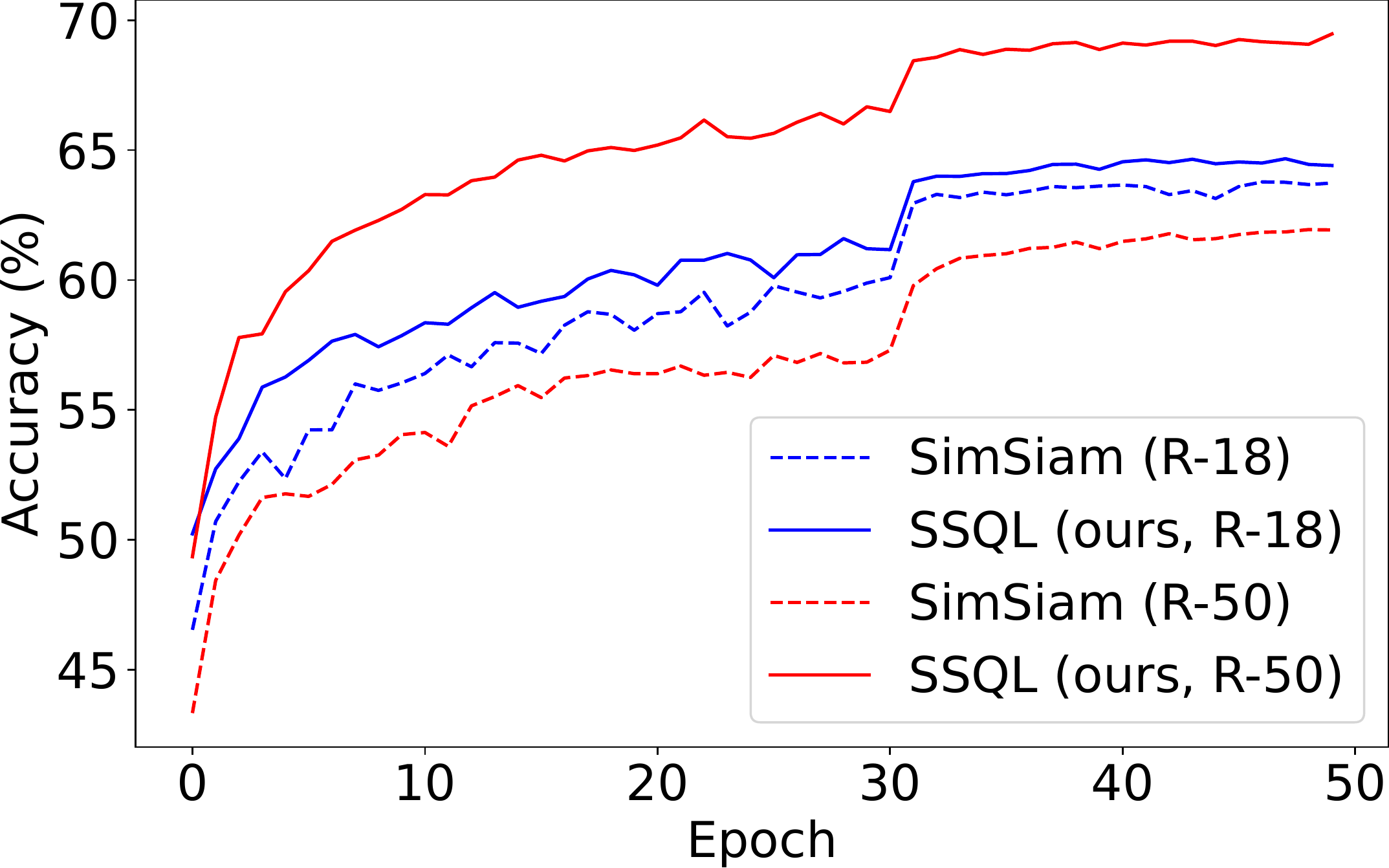}
	}
    \subfloat[2w4a]{
        \label{fig:LSQ-2w4a}
        \includegraphics[width=0.3\linewidth]{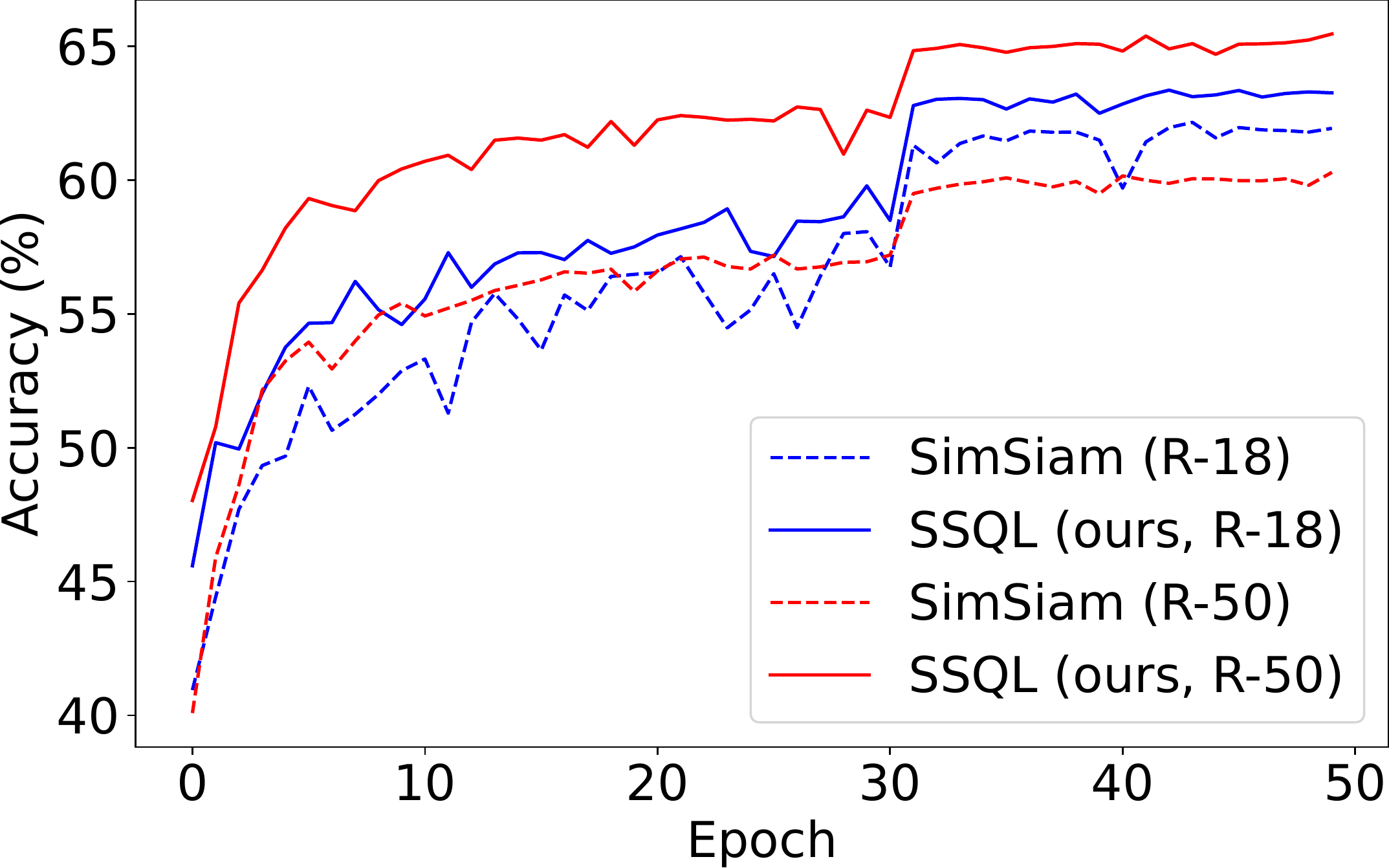}
	}
    \caption{ImageNet results using LSQ~\cite{lsq:ICLR20}, initialized with the linear evaluation models in Table~\ref{tab:imagenet_linear}. See appendix for details of the settings for LSQ.}
    \label{fig:LSQ-imagenet}
\end{figure}

\noindent\textbf{Combining with QAT method.} To further illustrate the effectiveness of our method, we combine different pretrained models with the state-of-the-art QAT method LSQ~\cite{lsq:ICLR20}. We initialize LSQ with ImageNet linear evaluated FP models (i.e., FP column in Table~\ref{tab:imagenet_linear}), where the backbone weights are learned by SSL methods and the linear classifier is fine-tuned with labels. The learning curves are shown in Fig.~\ref{fig:LSQ-imagenet} and we can see that our SSQL provides a better starting point for low-bit QAT training. Take R-50 4w4a as an example, SSQL achieves 7\% higher accuracy than SimSiam (54.2 v.s. 47.2) after the first epoch, while the initial accuracy of the FP model is about the same (67.9 v.s. 68.1). Consequently, our SSQL achieves higher final accuracy and it shows that our pretrained model can serve as a better initialization when combined with QAT methods to boost performance. We also combine with LSQ after fine-tuning different pretrained models on COCO2017 and observe similar improvements in the appendix.

\begin{table}[t]
	\caption{ImageNet transfer results on recognition benchmarks under R-50.}
	\label{tab:transfer_classification_r50}
	\centering
	\renewcommand{\arraystretch}{0.85}
	\renewcommand{\multirowsetup}{\centering}
	\begin{tabular}{c|c|c|c|c|c|c|c|c|c|c|c}
		\hline
		\multirow{2}{*}{Datasets}  & \multirow{2}{*}{Method} & \multicolumn{5}{c|}{Linear evaluation} & \multicolumn{5}{c}{Fine-tuning} \\
		\cline{3-12}
		&  & FP & 8w8a & 5w5a & 4w4a & 3w3a  & FP & 8w8a & 5w5a & 4w4a & 3w3a  \\
		\hline
		\multirow{2}{*}{CIFAR-10} & SimSiam &86.3 &86.2 &84.4 & 70.7 & 48.0   & 95.9	&95.9 &  92.0 &  51.7 & 14.6   \\
		&  SSQL (ours) & \textbf{89.3} & \textbf{89.2} & \textbf{89.1} & \textbf{87.1} & \textbf{71.9}  & \textbf{96.3} & \textbf{96.3} & \textbf{95.1} & \textbf{89.2} & \textbf{69.3}   \\
		\hline
		\multirow{2}{*}{CIFAR-100} & SimSiam &58.9& 58.7& 52.5 & 39.0 & 20.2   & 82.9 & 82.5 & 76.7 & 66.0 & \phantom{0}5.3    \\
		&  SSQL (ours) & \textbf{68.7} & \textbf{68.6} & \textbf{68.8} &  \textbf{66.4} & \textbf{49.7}  & \textbf{83.3} & \textbf{83.3}& \textbf{82.0}& \textbf{74.7} & \textbf{39.3}  
		\\
		\hline
		\multirow{2}{*}{Flowers} & SimSiam & 78.7 & 82.5 & 81.9 & 66.6 & 49.3  &94.0	&93.8 & 83.8& 57.8 & 16.2   \\
		&  SSQL (ours) & \textbf{90.7} & \textbf{90.7} &\textbf{91.3} & \textbf{90.9} & \textbf{84.0}  & \textbf{95.3} & \textbf{95.3} & \textbf{94.6}& \textbf{90.3} & \textbf{70.6}     \\
		\hline 
		\multirow{2}{*}{Food-101} & SimSiam & 67.1 & 67.1 & 64.7 & 56.0 & 27.7   & \textbf{86.2} & \textbf{86.2} & 80.4 & 54.4 & \phantom{0}2.2   \\
		&  SSQL (ours) & \textbf{72.6} & \textbf{72.5} & \textbf{71.5} & \textbf{68.4} & \textbf{51.6}   & 85.5 & 85.5 & \textbf{84.5}& \textbf{70.4} & \textbf{11.2}    \\
		\hline 
		\multirow{2}{*}{Pets} & SimSiam &79.7 & 79.6 &  74.3 & 70.9 & 32.2 & \textbf{87.5} & \textbf{87.4} & 81.3 & 59.3 & 10.9    \\
		&  SSQL (ours) & \textbf{83.6} & \textbf{83.9} &\textbf{83.3} & \textbf{82.3} & \textbf{73.8}  & 86.9 & 86.8 & \textbf{85.9} & \textbf{84.6} & \textbf{73.6}   \\
		\hline 
		\multirow{2}{*}{Dtd} & SimSiam & 69.9 & 69.7 & 69.1 & 63.4 & 46.6    & 73.4 & 73.6 & 70.5 & 60.4 & \phantom{0}8.8  \\
		&  SSQL (ours) & \textbf{74.4} & \textbf{74.3} &\textbf{74.3} &  \textbf{73.4} & \textbf{64.4}   & \textbf{73.7} &	\textbf{73.7} & \textbf{71.9} & \textbf{70.1} & \textbf{56.6}  \\
		\hline 
		\multirow{2}{*}{Caltech-101} & SimSiam & 80.2 & 80.4& 78.6 & 66.7 & 31.4   & \textbf{86.9} & \textbf{86.6} & 85.0 & 76.8 & \phantom{0}7.9   \\
		&  SSQL (ours) &\textbf{86.9} & \textbf{87.2} & \textbf{85.2} & \textbf{83.8} & \textbf{65.9}   & 86.4	& 86.3&\textbf{85.5}& \textbf{82.9} & \textbf{59.7}   \\
		\hline 
	\end{tabular}
\end{table}

\begin{figure}[t]
    \centering
    \subfloat[CIFAR-10+Lin]{
        \label{fig:cifar10-linear}
        \includegraphics[width=0.23\linewidth]{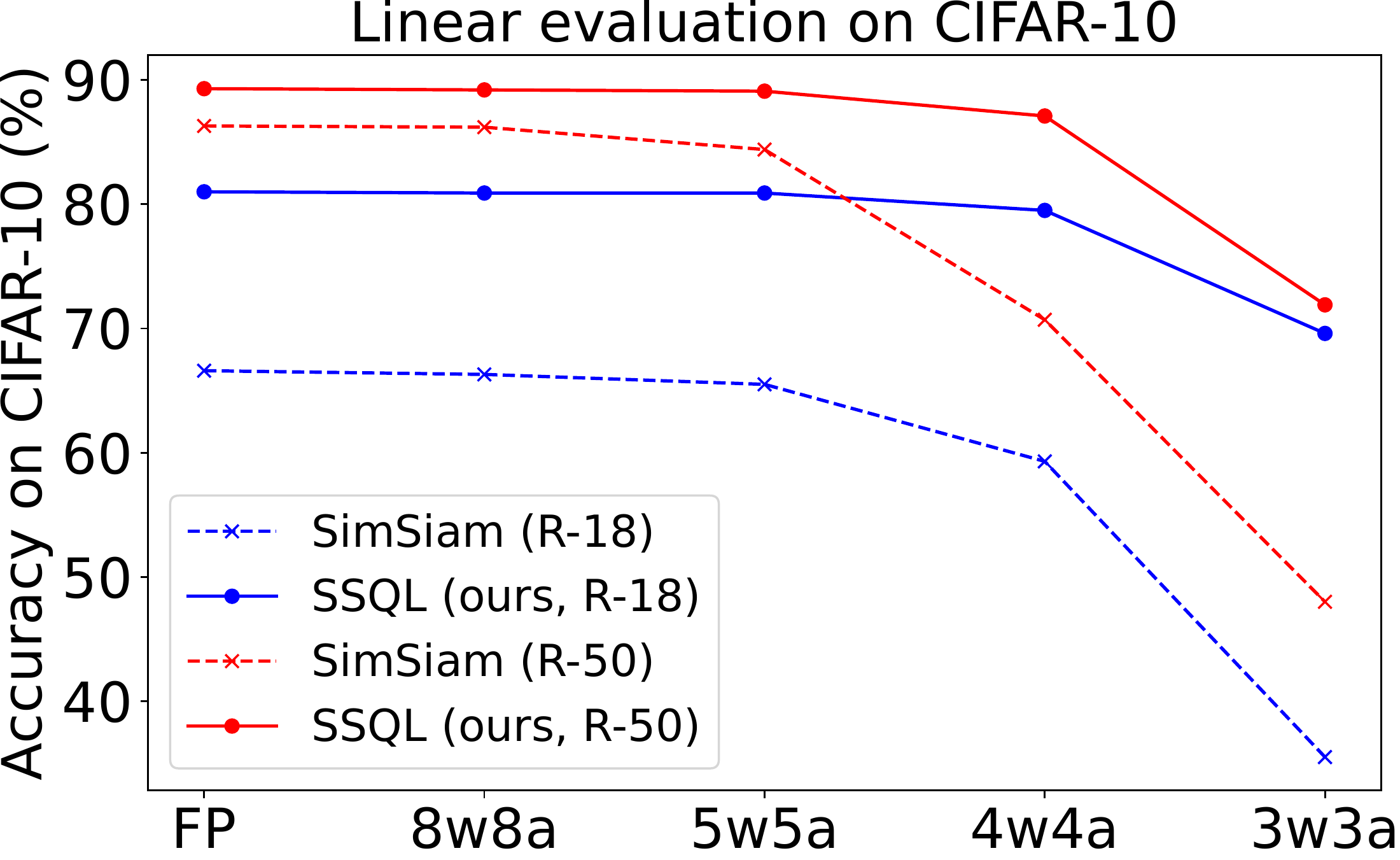}
	}
    \subfloat[Flowers+Lin]{
        \label{fig:flowers-linear}
        \includegraphics[width=0.23\linewidth]{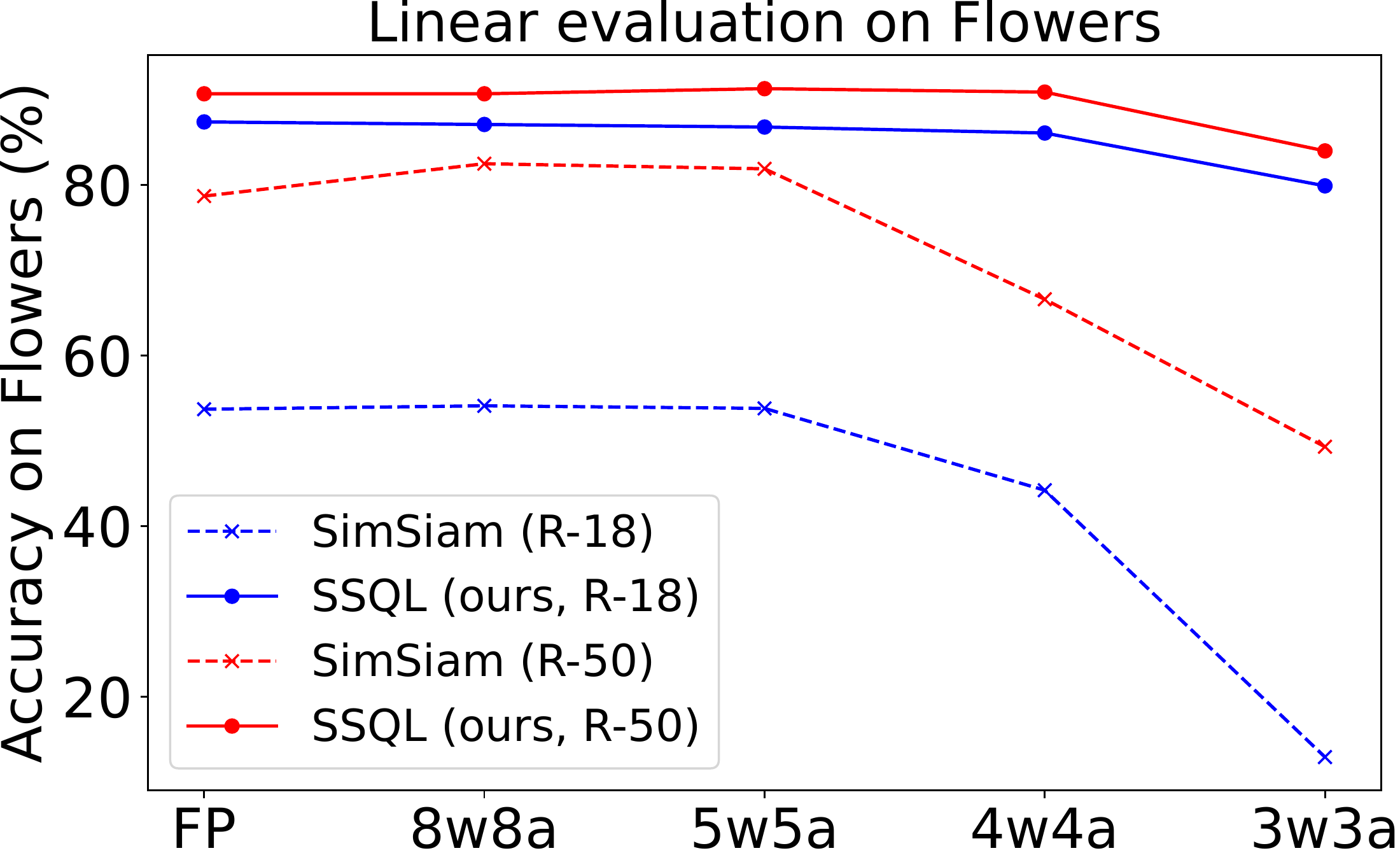}
	}
    \subfloat[Dtd+Lin]{
        \label{fig:dtd-linear}
        \includegraphics[width=0.23\linewidth]{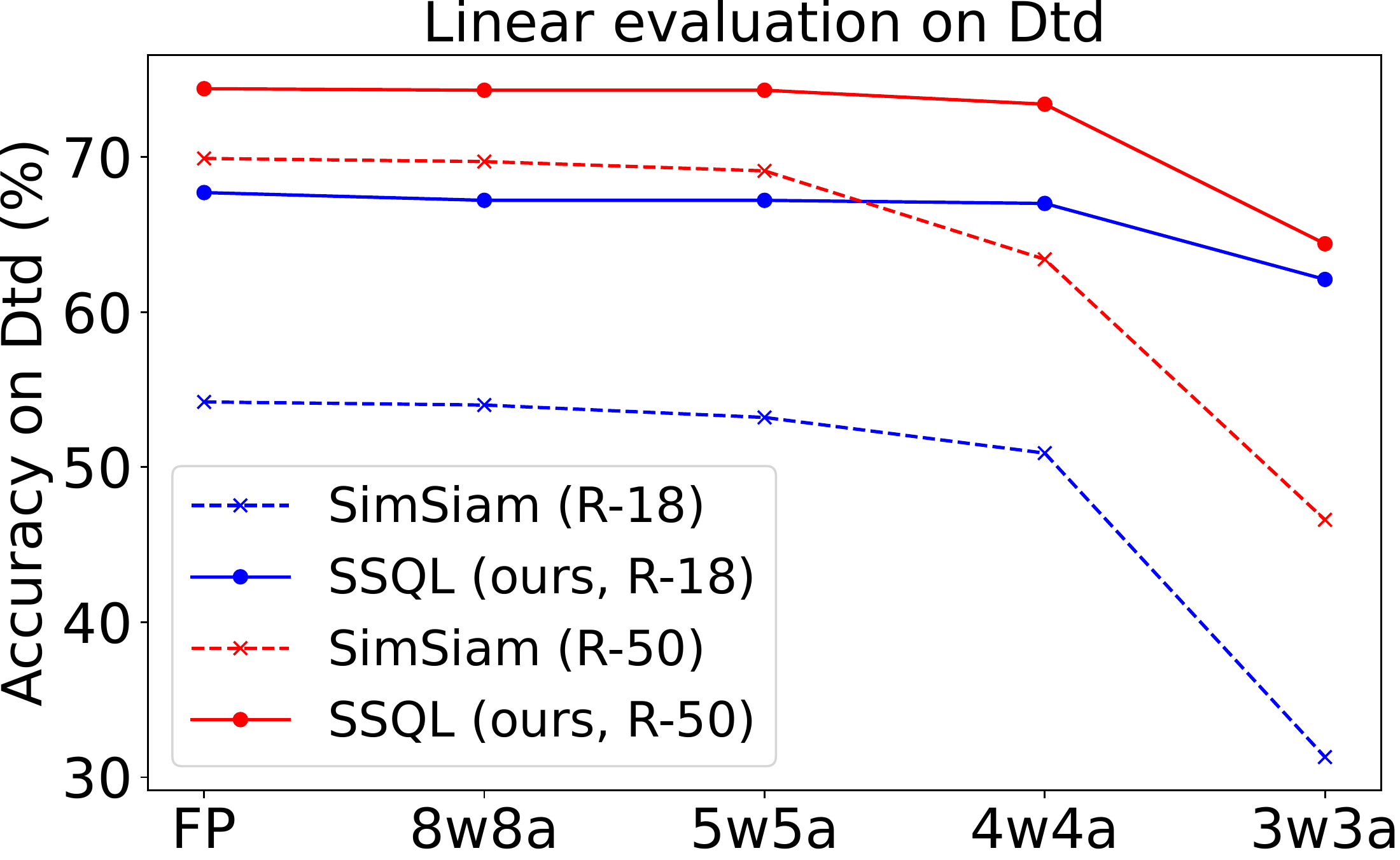}
    }
    \subfloat[Caltech-101+Lin]{
        \label{fig:caltech-linear}
        \includegraphics[width=0.23\linewidth]{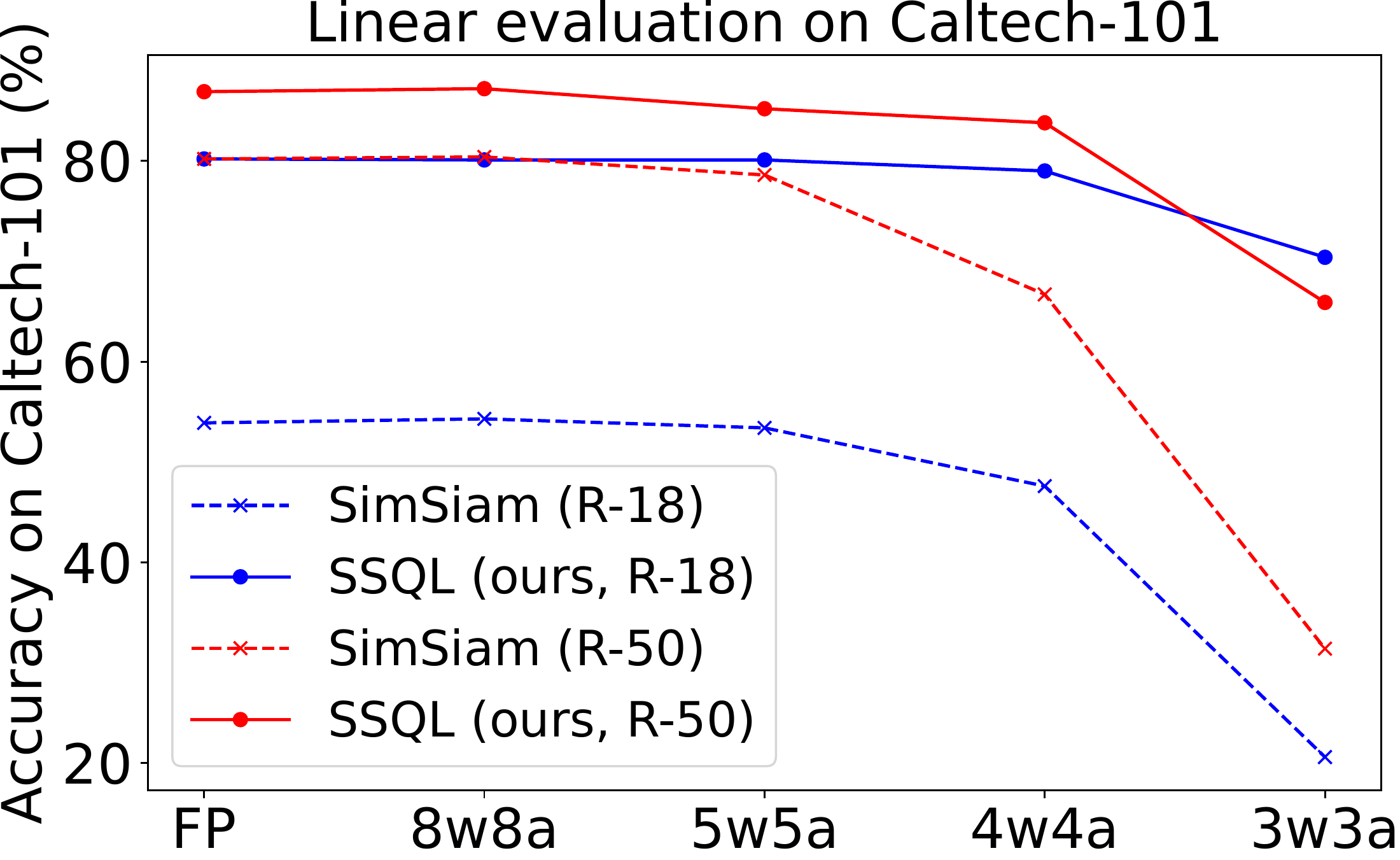}
    }
    \\
    \subfloat[CIFAR-10+FT]{
        \label{fig:cifar10-finetune}
        \includegraphics[width=0.23\linewidth]{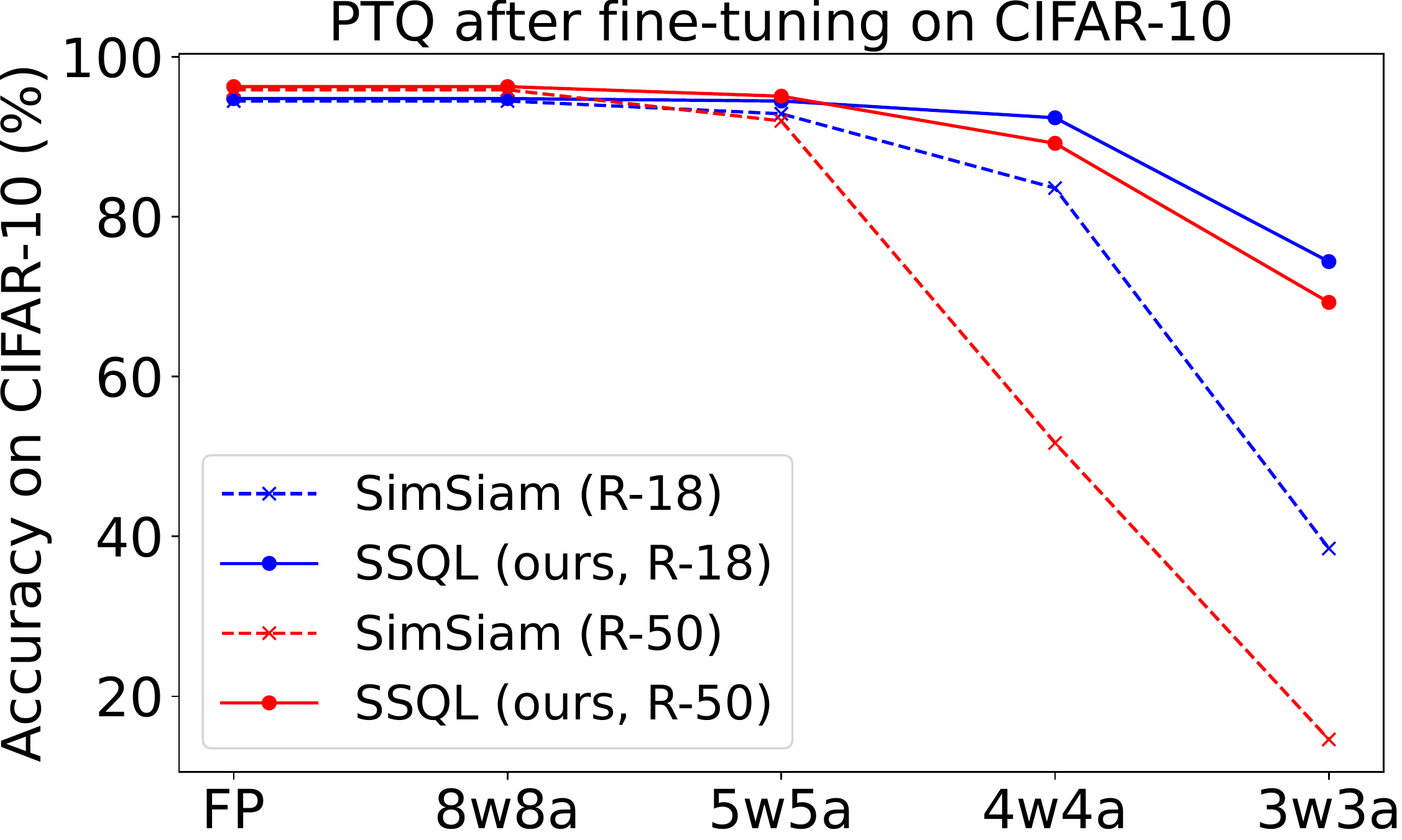}
	}
    \subfloat[Flowers+FT]{
        \label{fig:flowers-finetune}
        \includegraphics[width=0.23\linewidth]{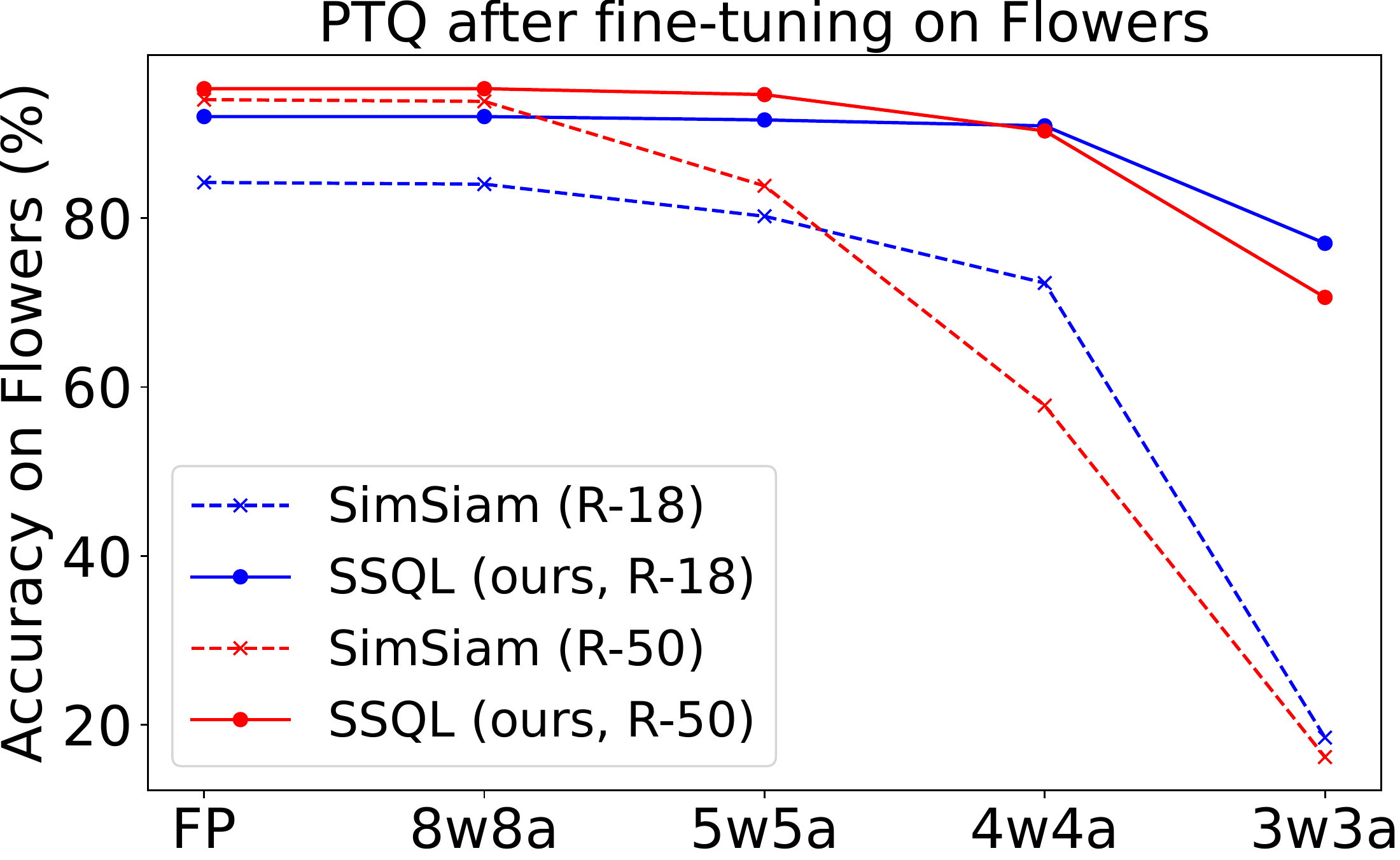}
	}
    \subfloat[Dtd+FT]{
        \label{fig:dtd-finetune}
        \includegraphics[width=0.23\linewidth]{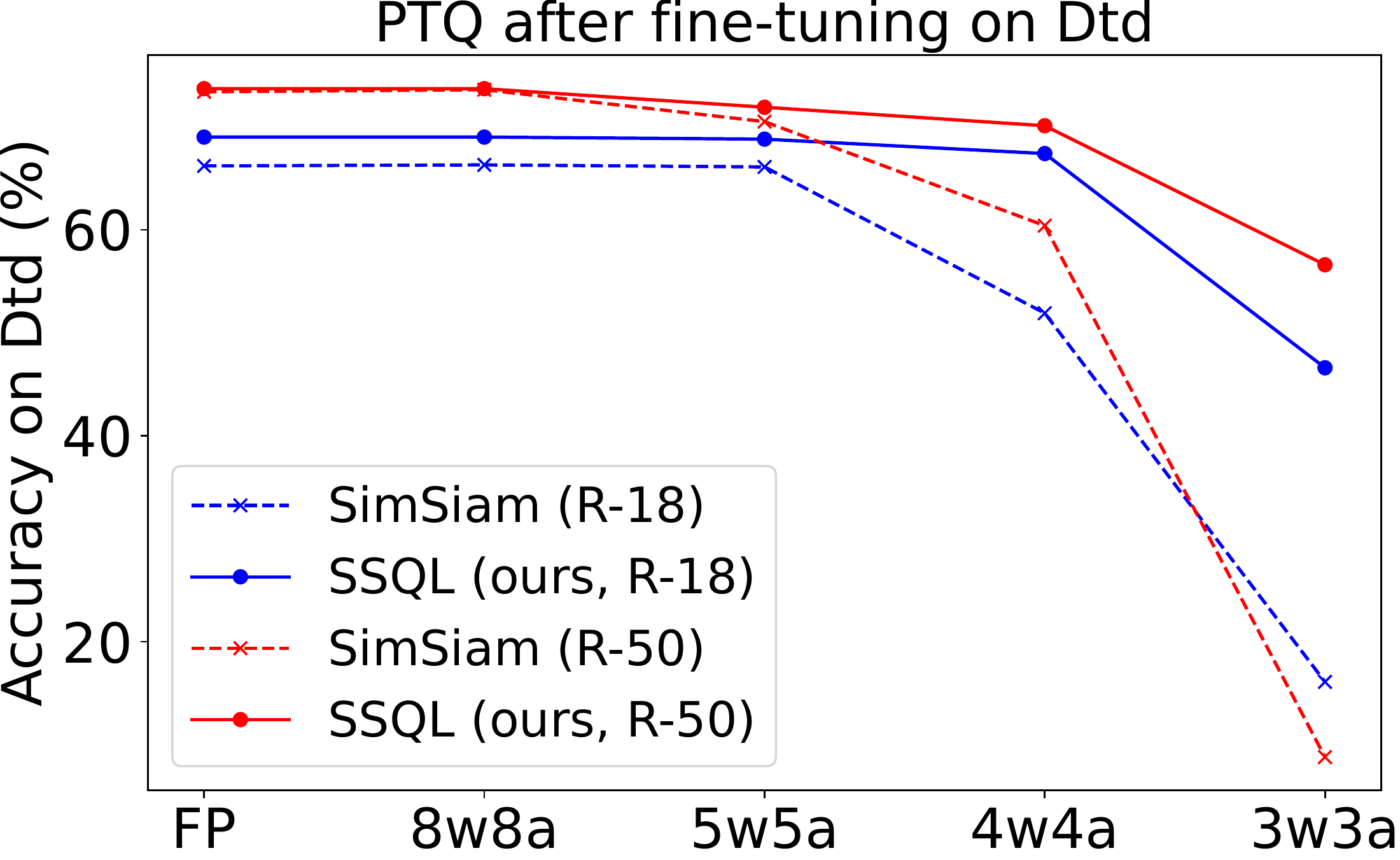}
    }
    \subfloat[Caltech-101+FT]{
        \label{fig:caltech-finetune}
        \includegraphics[width=0.23\linewidth]{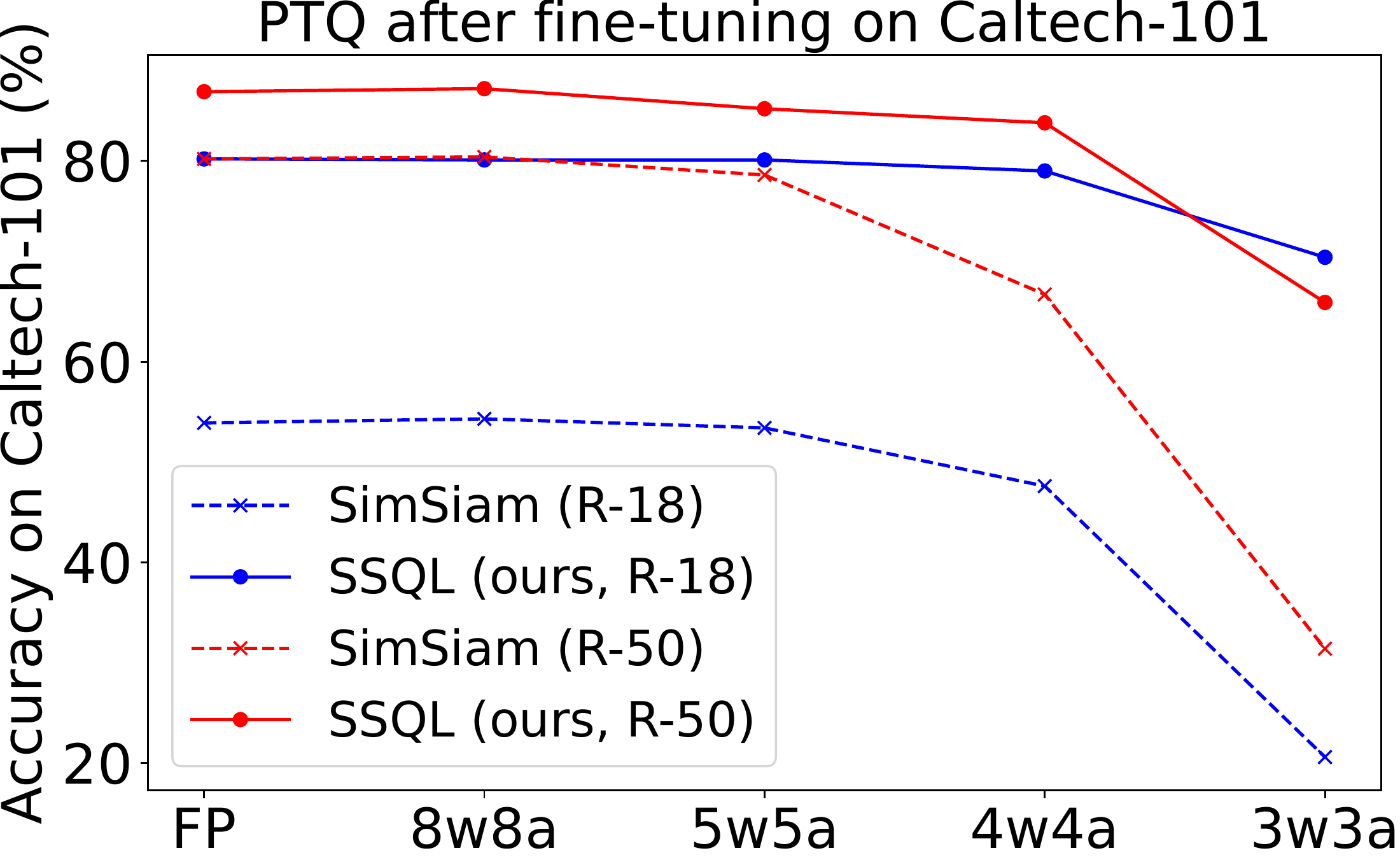}
    }
    \caption{Transfer recognition results. The first and second row shows the results under linear evaluation and fine-tuning (`FT'), respectively. Best viewed in color.}
    \label{fig:transfer-classification}
\end{figure}

\noindent\textbf{Transferring to recognition benchmarks.} We transfer the ImageNet learned representations of R-50 to downstream recognition tasks in Table~\ref{tab:transfer_classification_r50}. The training details and evaluation protocols are included in Sec.~\ref{sec:details} and the results of R-18 are included in the appendix. As shown in Table~\ref{tab:transfer_classification_r50}, our method improves a lot on all recognition benchmarks, especially under linear evaluation. When considering linear evaluation (backbone weights frozen), we can see that our SSQL has better transfer ability than SimSiam and the quantization-friendly properties are also well-preserved when transferring to downstream datasets.

When comparing the fine-tuning results at FP, we can see that our SSQL achieves comparable results under R-50. When we further conduct PTQ, we can observe larger improvements as the bit-width decreases, which is consistent with the properties observed in upstream pretraining. Take R-50 on CIFAR-10 as an example, SSQL is slightly better than SimSiam at FP but the improvement expands to 37.5\% at 4w4a and 54.7\% at 3w3a. We also plot the results in Fig.~\ref{fig:transfer-classification} and we can more clearly see the advantages of our SSQL (solid v.s. dotted line). In conclusion, the quantization-friendly properties are also well-preserved by SSQL when we fine-tune the weights during transferring. This again confirms our motivation that quantization-friendly pretraining is both important and feasible.

\noindent\textbf{Transferring to object detection.} We investigate the downstream object detection performance on Pascal VOC07\&12~\cite{VOC:mark:IJCV10} in Table~\ref{tab:transfer_voc_ptq} and COCO2017~\cite{coco:LinTY:ECCV14} in Table~\ref{tab:transfer_coco_ptq}. The detector is Faster R-CNN~\cite{faster-rcnn:ren:NIPS15} with a backbone of R18-C4~\cite{mask-rcnn:he:ICCV17} for VOC and Mask R-CNN~\cite{mask-rcnn:he:ICCV17} with R50-FPN~\cite{FPN:kaiming:CVPR17} backbone for COCO, implemented in \cite{wu2019detectron2}. We follow the same settings in~\cite{mocov2:xinlei:arxiv2020} and we evaluate the performance of post-training quantization models (i.e., the fine-tuning+PTQ pipeline).  

\begin{table}[t]
	\caption{Object detection results on VOC2007 under R18-C4. The best results are in \textbf{boldface} and the second best results are \underline{underlined}.}
	\label{tab:transfer_voc_ptq}
	\centering
	\resizebox{\linewidth}{!}{
	\setlength{\tabcolsep}{1pt}
	\renewcommand{\arraystretch}{0.9}
	\renewcommand{\multirowsetup}{\centering}
	\begin{tabular}{c|c|c|c|c|c|c|c|c|c|c|c|c|c|c|c}
	    \hline
		\multirow{2}{*}{Method} & \multicolumn{3}{c|}{FP} & \multicolumn{3}{c|}{8w8a} & \multicolumn{3}{c|}{6w6a} & \multicolumn{3}{c|}{5w5a}& \multicolumn{3}{c}{4w4a} \\
		\cline{2-16}
	    & $\text{AP}_{50}$ & $\text{AP}$ & $\text{AP}_{75}$ & $\text{AP}_{50}$ & $\text{AP}$ & $\text{AP}_{75}$& $\text{AP}_{50}$ & $\text{AP}$ & $\text{AP}_{75}$& $\text{AP}_{50}$ & $\text{AP}$ & $\text{AP}_{75}$& $\text{AP}_{50}$ & $\text{AP}$ & $\text{AP}_{75}$\\
		\hline
		random init. & 58.9 & 32.1 & 30.5 & 58.7 & 31.9 & 30.2 &  58.4 & 31.6 & 30.2 & 57.0 & 30.4 & 28.9 & 42.4 & 20.8 & 17.0 \\
		IN supervised & \textbf{73.9} & 44.6 & 46.5 & \textbf{74.1} & 44.2 & 46.2 & 73.0 & 43.4 & 44.5 & 68.9 & 39.4 & 39.3 & 33.1 & 16.7 & 14.0 \\
		BYOL & 72.8 & 44.7 & 46.3 & 72.4 & 44.4 & 46.0 & 72.2 & 44.2 & 45.6 & 62.7 & 38.0 & 39.4 & 52.7 & 28.9 & 27.8\\
		SimSiam & 72.8 & 44.4 & 46.6 & 72.9 & 44.4 & 46.3 & 72.4 & 44.0 & 46.0 & 69.7 & 42.0 & 43.1 & 50.4 & 26.4 & 23.8 \\
		SimSiam-200ep & 72.5 &44.3 & 46.5 & 72.5 & 44.3 & 46.5 & 72.0 & 43.9 & 46.3 & 69.2 & 41.4 & 42.7 & 53.7 & 29.8 & 29.0 \\
		SSQL (ours) & \underline{73.4} & \underline{44.7} & \underline{46.8} & \underline{73.5} & \textbf{45.0} & \underline{46.8} & \textbf{73.1} &  \underline{44.5}  & \underline{46.4} & \textbf{71.6} & \underline{42.8} & \underline{44.4} & \textbf{61.2} & \underline{34.1} & \underline{33.4} \\
		SSQL-200ep (ours) & 73.2 & \textbf{45.0} & \textbf{47.3} & 73.2&\textbf{45.0}&\textbf{47.0}&\underline{72.9}&\textbf{44.8}&\textbf{46.8}&\underline{71.3}&\textbf{43.3}&\textbf{45.0}& \textbf{61.2} & \textbf{35.1} & \textbf{35.0}  \\
		\hline 
	\end{tabular}
   }
\end{table}

\begin{table}[t]
	\caption{Object detection/segmentation results on COCO2017 under R50-FPN.}
	\label{tab:transfer_coco_ptq}
	\centering
	\resizebox{\linewidth}{!}{
	\setlength{\tabcolsep}{1pt}
	\renewcommand{\arraystretch}{0.85}
	\renewcommand{\multirowsetup}{\centering}
	\begin{tabular}{c|c|c|c|c|c|c|c|c|c|c|c|c}
		\hline
		\multirow{2}{*}{Method} & \multicolumn{6}{c|}{FP} & \multicolumn{6}{c}{6w6a}   \\
		\cline{2-13}
	     &$\text{AP}^{\text{bb}}$  &  $\text{AP}_{50}^{\text{bb}}$  &  $\text{AP}_{75}^{\text{bb}}$  &  $\text{AP}^{\text{mk}}$  &  $\text{AP}_{50}^{\text{mk}}$  & $\text{AP}_{75}^{\text{mk}}$ & $\text{AP}^{\text{bb}}$  &  $\text{AP}_{50}^{\text{bb}}$  &  $\text{AP}_{75}^{\text{bb}}$  &  $\text{AP}^{\text{mk}}$  &  $\text{AP}_{50}^{\text{mk}}$  & $\text{AP}_{75}^{\text{mk}}$\\
		\hline
		IN supervised & 38.2 & 56.0 & 42.0 & 34.8 & 56.0 & 37.2 & 37.6 & 58.3 & 41.4 &  34.3 &  55.2 & 36.8 \\
		SimSiam & \textbf{38.9} & \textbf{59.8} & \textbf{42.3} &  \textbf{35.2} & \textbf{56.7} & \textbf{37.7} & 38.1 & 58.7 & 41.5 & 34.5 & 55.7 & 36.8 \\
		BYOL & 37.4	& 57.9 & 40.6 & 34.1 & 54.9 & 36.4 & 37.0 & 57.4 & 40.2 & 33.7 & 54.3&36.0 \\
		SSQL (ours) & 38.7 & 59.2 & \textbf{42.3} & \textbf{35.2} & 56.2 & \textbf{37.7} & \textbf{38.3} & \textbf{58.8} & \textbf{41.7} & \textbf{34.8} & \textbf{55.8} & \textbf{37.3}  \\
		\hline 
		&  \multicolumn{6}{c|}{5w5a} & \multicolumn{6}{c}{4w4a} \\
	    \cline{1-13}
	    IN supervised & 35.2 &55.5 & 38.4 & 31.9 & 52.3 & 34.0 & 23.4 & 38.6 & 24.6 & 21.4 &  36.3 & 22.1\\
	    SimSiam & 34.3 & 54.0 & 36.7 & 30.9 & 50.6 & 32.6 & 19.9 & 33.6 & 20.6 & 18.1 & 31.3 & 18.3  \\
	    BYOL &  34.9& 54.4&37.7&31.8&51.4&33.8 & 22.7 & 37.4 & 24.0 & 20.9 & 35.2 & 21.7 \\
		SSQL (ours)& \textbf{36.5} & \textbf{56.9} & \textbf{39.4} & \textbf{33.3} & \textbf{53.6} & \textbf{35.5} & \textbf{28.2} & \textbf{43.1} & \textbf{27.5} & \textbf{26.0} & \textbf{43.1} & \textbf{27.5}   \\
		\hline
	\end{tabular}
   }
\end{table}

As shown in Table~\ref{tab:transfer_voc_ptq}, our SSQL performs better than SimSiam and BYOL on Pascal VOC at FP. Also, as we lower the bit-width, our SSQL is more significantly better than baseline counterparts: up to \textbf{+1.9} and \textbf{+7.5} $\text{AP}_{50}$ over \textit{the best results among other methods} at 5w5a and 4w4a, respectively. We can reach similar conclusions on COCO2017 from Table~\ref{tab:transfer_coco_ptq}. Although our SSQL achieves slightly lower accuracy than SimSiam at FP on COCO, we achieve \textbf{+2.2} and \textbf{+8.3} $\text{AP}^\text{bb}$ points higher at 5w5a and 4w4a, respectively. In conclusion, the results show that the quantization-friendly property of our pretrained model can be well-preserved even after fine-tuning on downstream detection tasks. 

\begin{table}[t]
	\caption{Ablation studies on CIFAR-10 using ResNet-34.}
	\label{tab:ablation-R34}
 	\centering
 	\renewcommand{\arraystretch}{0.85}
  	\renewcommand{\multirowsetup}{\centering}
	\begin{tabular}{c|c|c|c|c|c|c|c|c|c|c|c}
		\hline
		\multirow{2}{*}{ID} & \multirow{2}{*}{Q Pred} &\multirow{2}{*}{Q Target}
		&\multirow{2}{*}{Aux}
		&\multirow{2}{*}{W Bit} & \multirow{2}{*}{A Bit} & \multicolumn{6}{c}{Linear evaluation accuracy (\%)} \\
		\cline{7-12}
		& & && && FP & 6w6a & 4w4a & 3w3a & 2w4a & Avg.\\
		\hline
		(a)& $\times$  & $\times$ &$\times$& -& - & 89.0 & 89.0 & 87.2 & 75.6 & 55.3 & 79.2\\
		(b)&$\times$ & $\checkmark$ &$\times$& 4$\sim$16 & 4$\sim$16 & 87.6  & 87.5 & 85.8 & 70.4 & 58.5 & 78.0 \\
		(c)&$\checkmark$ & $\checkmark$ &$\times$& 4$\sim$16 & 4$\sim$16 &  90.5 & 90.4 & 88.9 & 79.2 & 73.7 & 84.5 \\
		(d)& $\checkmark$ & $\times$ & $\times$& 4$\sim$16 & 4$\sim$16 &  \textbf{91.0} & \textbf{91.0} & 89.5 & 83.0 & 65.2 & 83.9 \\
		(e)&$\checkmark$ & $\times$ & $\times$& 6 & 6 & 90.0&89.9&87.9 & 69.1 & 62.1 & 79.8 \\
		(f)&$\checkmark$ & $\times$ & $\times$& 4 & 4 & 36.0&35.9&36.4&29.2&29.7 & 33.4 \\
		(g)&$\checkmark$ & $\checkmark$ &$\times$& 2$\sim$8 & 4$\sim$8 & 88.3  & 88.2 & 86.9 & 80.3 & 85.4 & 85.8 \\
		(h)&$\checkmark$ & $\times$ &$\times$& 2$\sim$8 & 4$\sim$8 &   89.6 & 89.5	& 88.2 & 82.9 & 81.5 & 86.3 \\
		(i)&$\checkmark$ & $\times$ &$\checkmark$& 2$\sim$8 & 4$\sim$8 &  90.9 & 90.8 & \textbf{89.6} & \textbf{83.2} & \textbf{86.8} & \textbf{88.3}\\
		\hline
	\end{tabular}
\end{table}

\subsection{Ablation studies} \label{sec:ablation}
We conduct ablation studies on CIFAR-10 under ResNet-34 and we keep the training settings the same as in Sec.~\ref{sec:exp-cifar}, as shown in Table~\ref{tab:ablation-R34}. `Q Pred' denotes whether to quantize the prediction branch and the same for `Q Target'. `Aux' denotes whether to add the auxiliary SimSiam loss. `W/A Bit' represents the candidate bit-widths set for weight/activation. We can have the following conclusions from Table~\ref{tab:ablation-R34}:

\squishlist
\item Quantizing the target branch only degenerates the performance. The row (b) is the worst among the first four rows, which indicates that only using the quantized output as the target makes training more difficult (learning noisy targets). In other words, it is essential to update the quantized branch with the gradients (both row (c) and (d) perform better than the baseline row (a)).

\item Random selection of bit-widths for training is better than training with a single bit-width. We can observe that the row (d) surpasses the row (e) and (f) at all bit-widths, where the latter two are trained using a single bit-width. It shows that the random selection operation in our method is beneficial to improve performance, by providing stronger randomness and augmentations.

\item Using a reasonable bit perturbation range further improves the performance at lower bit-widths. When comparing the row (d) and (h), we can observe a big boost at 2w4a (81.5 v.s. 65.2) at the expense of FP accuracy. When comparing the row (c) and (g), we can find that quantizing both branches at the same time results in a larger drop in FP accuracy. 

\item The row (i) achieves the best trade-off among all settings, which is also the default setting for all our experiments. When comparing the row (i) and (h), we can see that the addition of the auxiliary loss makes the full precision model produce better targets, thus improving the accuracies at all bit-widths. 
\squishend

\section{Conclusion}
In this paper, we proposed a method called SSQL for pretraining quantization-friendly models to facility flexible deployment in resource constrained applications. We provide theoretical analysis for the proposed approach, and experimental results on various benchmarks show that our method not only greatly improves the performance when quantized to lower bits, but also boosts the performance of full precision models. It has also been verified that our method is compatible with PTQ or QAT methods, and the quantization-friendly property can be well-preserved when transferring to downstream tasks. In the future, we will explore applications of SSQL to other architectures, notably Transformers. Also, we will explore fine-tuning methods that can better preserve the quantization-friendly property of our models.

\clearpage

\bibliographystyle{splncs04}
\bibliography{egbib}

\clearpage

\appendix

\section{Implementation details}

\noindent\textbf{Datasets.} The statistics of the classification benchmarks used in our paper are shown in Table~\ref{tab:dataset-overview}.

\begin{table}
	\caption{Statistics of the classification benchmarks used in the paper.}	\label{tab:dataset-overview}
	\centering
	\footnotesize
	\renewcommand{\multirowsetup}{\centering}
	\begin{tabular}{l|c|c|c}
			\hline
			Datasets & \# Category & \# Training & \# Testing \\
			\hline 
			CIFAR-10 &10 & 50,000 & 10,000\\
			CIFAR-100 &100 & 50,000 & 10,000\\
			Flowers &102& 2,040&6,149\\
			Food-101 & 101 & 75,750  & 25,250\\
			Pets &\pt37& 3,680&3,669\\
			DTD &\pt47& 3,760&1,880\\
			Caltech-101 & 101 & 2020 & 1010 \\
			\hline
	\end{tabular}
\end{table} 

\noindent\textbf{Training details for SSL methods.} Training details for MoCov2, SimCLR, BYOL, SimSiam and our SSQL on CIFAR-10/CIFAR-100 are shown in Table~\ref{tab:appendix-details}.

\begin{table}
	\caption{Training details for MoCov2, SimCLR, BYOL, SimSiam and our SSQL on CIFAR datasets in Table 1 and Table 2. $\tau$ denotes the temperature parameter, $k$ denotes the size of memory bank in MoCov2, and $m$ denotes the momentum in MoCov2 and BYOL.}
	\label{tab:appendix-details}
	\centering
	\small
	\setlength{\tabcolsep}{3.5pt}
	\begin{tabular}{l|c|c|c|c|c|c|c|c|c|c}
		\hline
		\multirow{2}{*}{Method}  & \multicolumn{10}{c}{Settings}  \\
		\cline{2-11}
		&bs&lr&wd&epochs&optimizer&lr schedule&$\tau$&$k$&$m$&dim\\
		\hline
		SimSiam & 512 & 0.05 &5e-4 &400&SGD & cosine & - & - & - & 2048 \\
		 MoCov2  &256&0.03& 1e-4 &400&SGD& cosine&0.2&4096& 0.999&2048\\ 
		 SimCLR   &512&0.5& 1e-4 & 400&SGD& cosine&0.5&-&-&2048\\ 
		 BYOL   &512&0.5& 5e-4 & 400&SGD& cosine&-&-&0.99&2048 \\ 
		SSQL (ours) & 256 & 0.05 & 1e-4 &400& SGD& cosine & - & - & -& 2048 \\
 		\hline
	\end{tabular}
\end{table}

\noindent\textbf{Training details for LSQ.} We initialize LSQ with linear evaluated full precision models on ImageNet. Then, we train 50 epochs using SGD. We set the batch size to 256, weight decay to 1e-5, and base lr to 0.001. We divide the learning rate by 10 at the 30th epoch.

\section{Experimental results}
We present more experimental results here in this section. We present more visualizations of weight distribution in Sec.~\ref{sec:app:distribution}, more transfer results in Sec.~\ref{sec:app:transfer}. Moreover, we investigate whether our method can be applied in more SSL frameworks in Sec.~\ref{sec:app:ssql_mocov2} and emerging new Transformer-like architectures in Sec.~\ref{sec:app:ssql_vit}.

\subsection{Weight distribution}\label{sec:app:distribution}
\begin{figure}[t]
	\centering
	\includegraphics[width=\columnwidth]{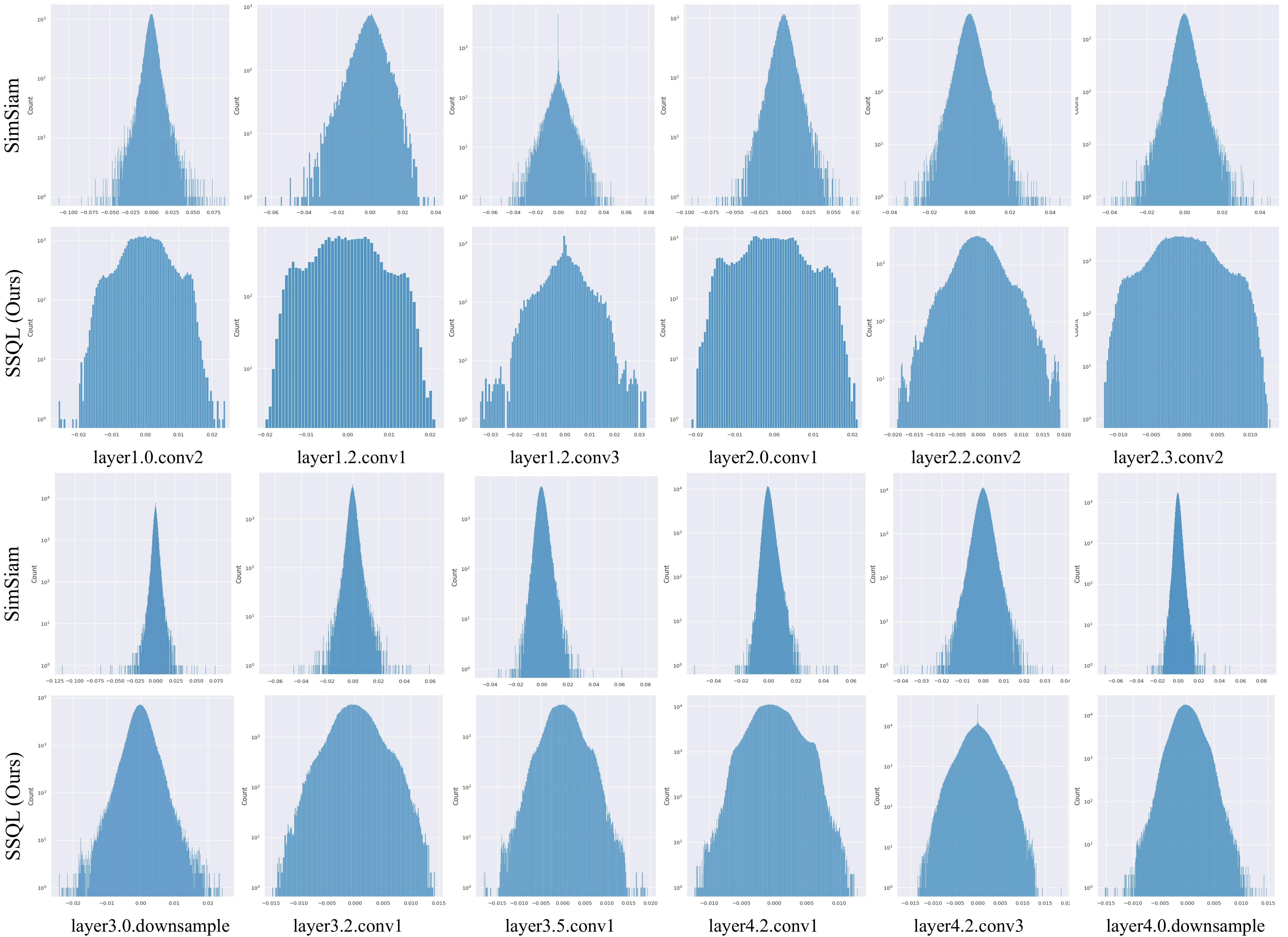}
	\caption{Visualization of weights distribution for ResNet-50 fine-tuned on Dtd.}
	\label{fig:weight-R-50-dtd}
\end{figure}

In figure~\ref{fig:weight-R-50-dtd}, we visualize the weights distribution of different models (fine-tuned from ImageNet pretrained models on Dtd). In this case, the backbone weights have been updated and we can still observe the quantization-friendly property of our model. As seen, our model has more uniform distribution, smaller ranges and much fewer outliers.

\begin{table*}[t]
	\caption{ImageNet transfer results on classification benchmarks under R-18.}
	\label{tab:transfer_classification_r18}
	\centering
	\renewcommand{\multirowsetup}{\centering}
	\begin{tabular}{c|c|c|c|c|c|c|c|c|c|c|c}
		\hline
		\multirow{2}{*}{Datasets}  & \multirow{2}{*}{Method} & \multicolumn{5}{c|}{Linear evaluation} & \multicolumn{5}{c}{Fine-tuning} \\
		\cline{3-12}
		&  & FP & 8w8a & 5w5a & 4w4a & 3w3a   & FP & 8w8a & 5w5a & 4w4a & 3w3a \\
		\hline
		\multirow{2}{*}{CIFAR-10} & SimSiam &66.6& 66.3& 65.5 & 59.3& 35.5   & 94.5 & 94.5 &  92.9 &  83.6 &  38.5 \\
		&  SSQL (ours) & \textbf{81.0} & \textbf{80.9} & \textbf{80.9} & \textbf{79.5} & \textbf{69.6}   & \textbf{94.8} & \textbf{94.8}&  \textbf{94.5} & \textbf{92.4} & \textbf{74.4}   \\
		\hline
		\multirow{2}{*}{CIFAR-100} & SimSiam & 33.2 & 33.2 & 32.3 & 25.3 & 10.9   & 77.0 & 77.0 & 73.7 & 56.0 & \phantom{0}9.5   \\
		&  SSQL (ours) & \textbf{55.8} & \textbf{55.9}&  \textbf{55.8} & \textbf{53.5} & \textbf{45.5}   & \textbf{79.0} & \textbf{78.8}&  \textbf{77.6} & \textbf{73.4} & \textbf{44.8}  
		\\
		\hline
		\multirow{2}{*}{Flowers} & SimSiam & 53.7&54.1& 53.8 &44.2& 12.9  & 84.2 & 84.0 & 80.2 & 72.3 & 18.5  \\
		&  SSQL (ours) & \textbf{87.4} & \textbf{87.1} & \textbf{86.8} & \textbf{86.1} & \textbf{79.9}   & \textbf{92.0} & \textbf{92.0} & \textbf{91.6}& \textbf{90.9} & \textbf{77.0}   \\
		\hline 
		\multirow{2}{*}{Food-101} & SimSiam & 36.4 & 35.0 &35.3 & 32.4 & 13.6   & \textbf{81.3} & \textbf{81.3} &  78.0 & 63.8 & 4.3  \\
		&  SSQL (ours) &  \textbf{60.7} & \textbf{60.9} &\textbf{59.9} & \textbf{57.6} & \textbf{48.7}  &  80.9  &  80.9  & \textbf{79.9} & \textbf{73.0} & \textbf{19.6}   \\
		\hline 
		\multirow{2}{*}{Pets} & SimSiam &48.3 & 48.7 & 47.7 & 42.4 & 6.2  & 80.2 &	80.2 & 77.9 & 54.1 & 8.8   \\
		&  SSQL (ours) &\textbf{77.3}& \textbf{77.3} & \textbf{77.0} & \textbf{75.1} & \textbf{66.3}   & \textbf{81.9} & \textbf{81.8} & \textbf{81.4} & \textbf{79.9} & \textbf{57.4}     \\
		\hline 
		\multirow{2}{*}{Dtd} & SimSiam & 54.2& 54.0 & 53.2 & 50.9 & 31.3   & 66.2 & 66.3 & 66.1 & 51.9 & 16.1   \\
		&  SSQL (ours) & \textbf{67.7} & \textbf{67.2} & \textbf{67.2} & \textbf{67.0} & \textbf{62.1} & \textbf{69.0} & \textbf{69.0} & \textbf{68.8}& \textbf{67.4} & \textbf{46.6}   \\
		\hline 
		\multirow{2}{*}{Caltech-101} & SimSiam & 53.9 & 54.3 & 53.4& 47.6 & 20.6   & 75.6 & 75.6 & 73.5 & 63.1 & 6.4   \\
		&  SSQL (ours) & \textbf{80.2} & \textbf{80.1} & \textbf{80.1} &  \textbf{79.0} & \textbf{70.4}   & \textbf{81.6} & \textbf{81.4} & \textbf{82.0} & \textbf{80.9} & \textbf{62.3}   \\
		\hline 
	\end{tabular}
\end{table*}

\begin{figure}[t]
    \centering
    \subfloat[$\text{AP}_{\text{bb}}$]{
        \label{fig:LSQ-coco-bb}
        \includegraphics[width=0.45\linewidth]{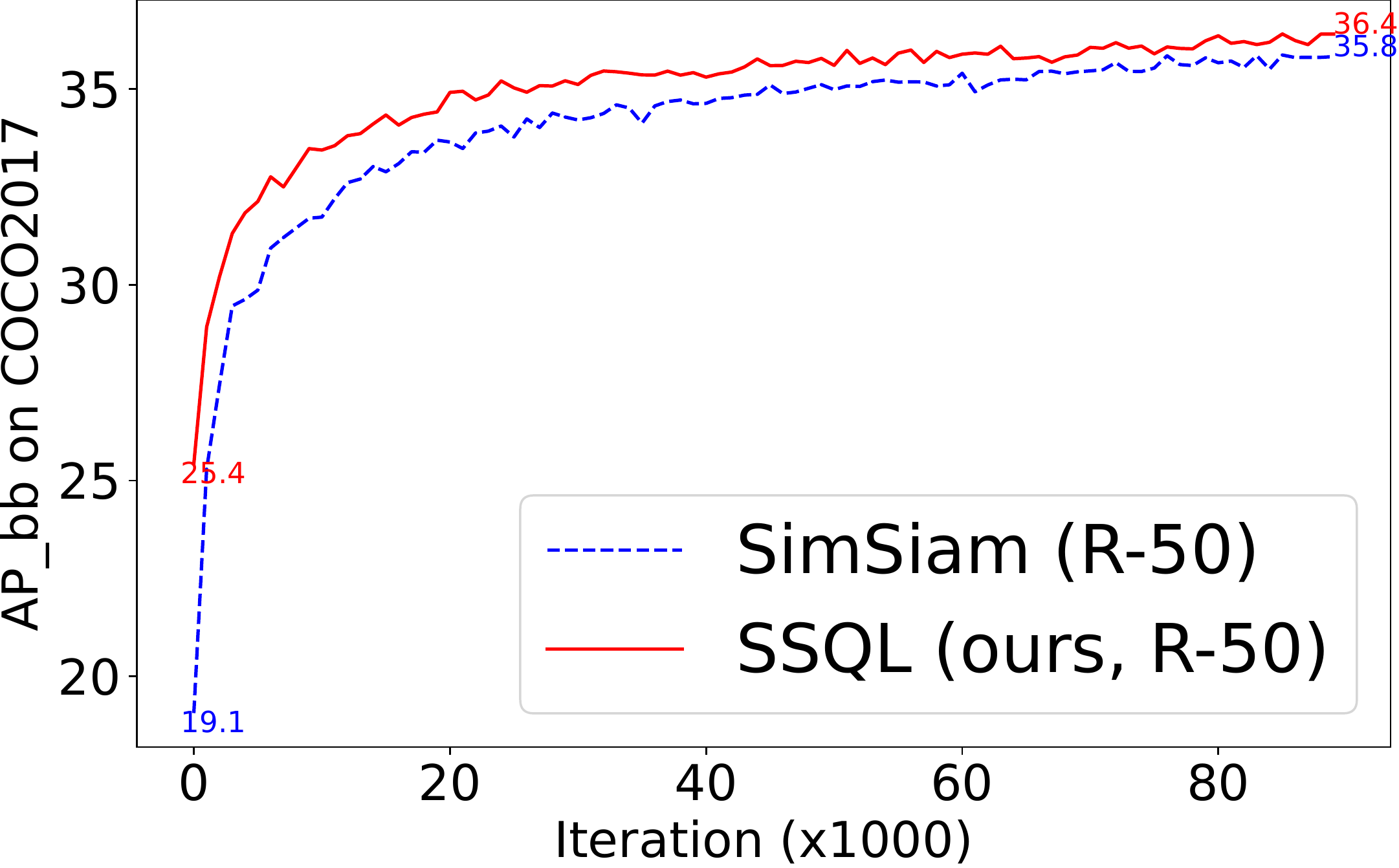}
	}
    \subfloat[$\text{AP}_{\text{mask}}$]{
        \label{fig:LSQ-coco-mask}
        \includegraphics[width=0.45\linewidth]{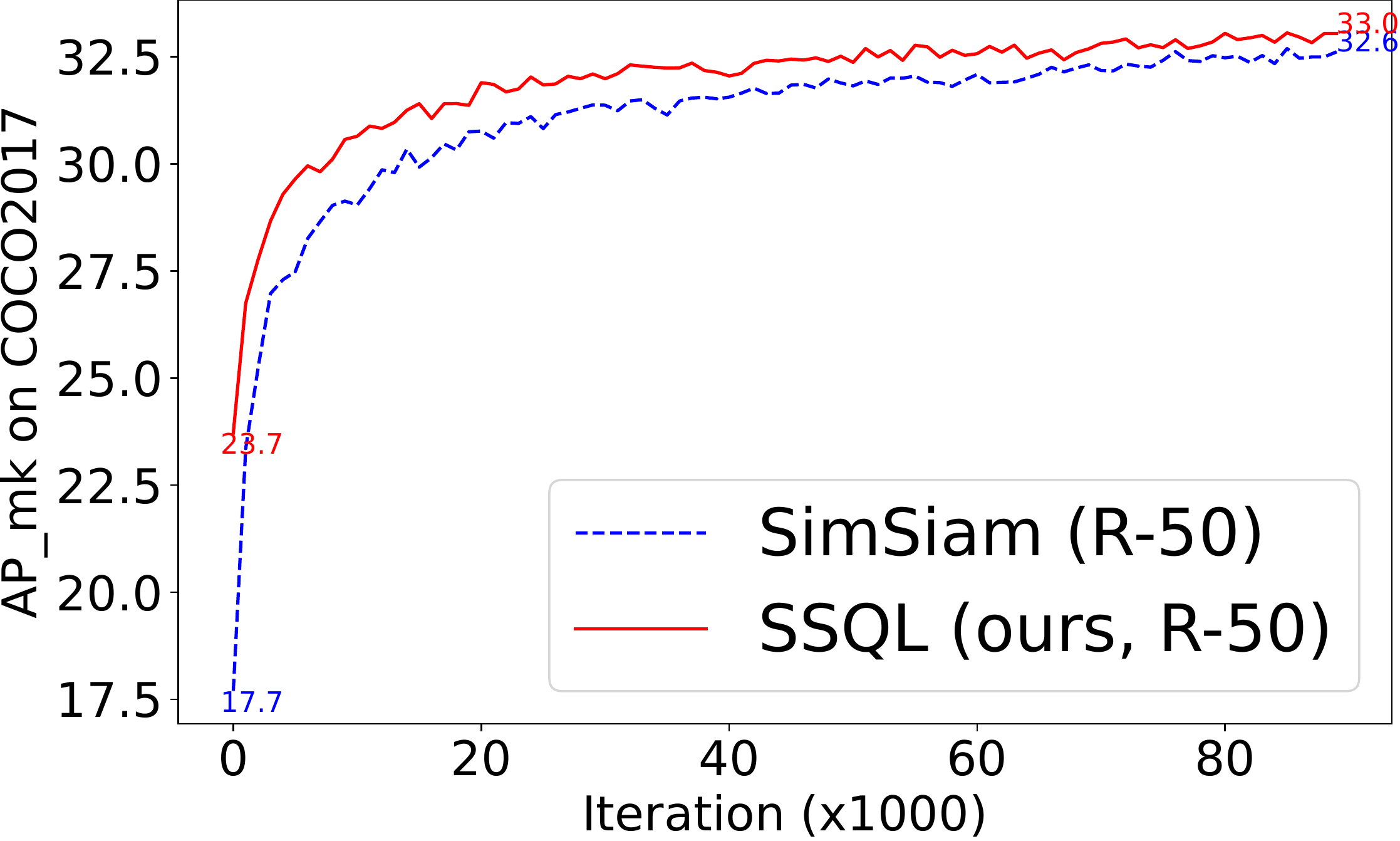}
	}
    \caption{COCO fine-tuning results using LSQ (2w4a), initialized with the fine-tuned FP models in Table~\ref{tab:transfer_coco_ptq}.}
    \label{fig:LSQ-coco}
\end{figure}

\begin{figure*}[t]
    \centering
    \subfloat[CIFAR-100+Lin]{
        \label{fig:appendix-cifar100-linear}
        \includegraphics[width=0.32\linewidth]{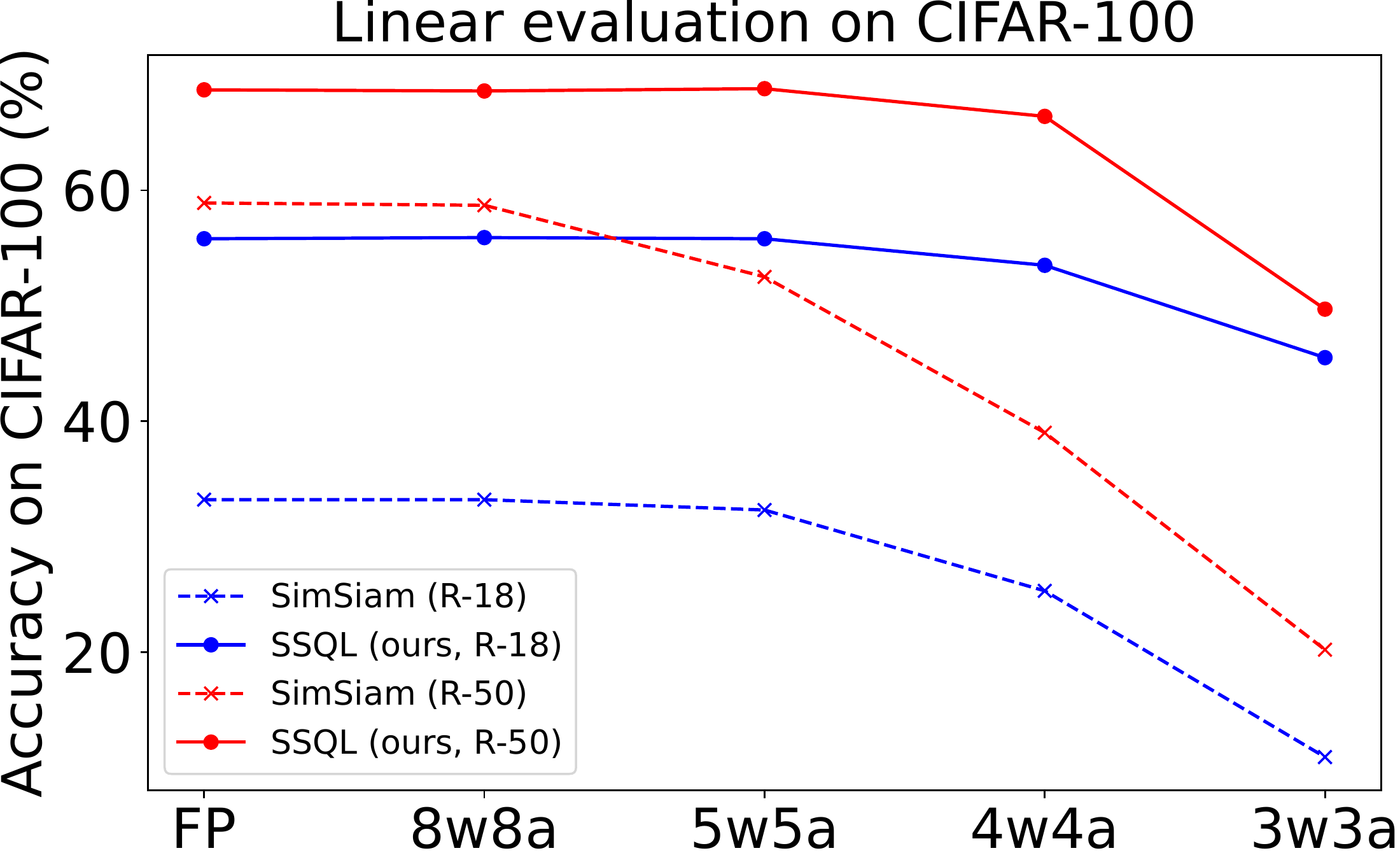}
	}
    \subfloat[Food-101+Lin]{
        \label{fig:appendix-food-linear}
        \includegraphics[width=0.32\linewidth]{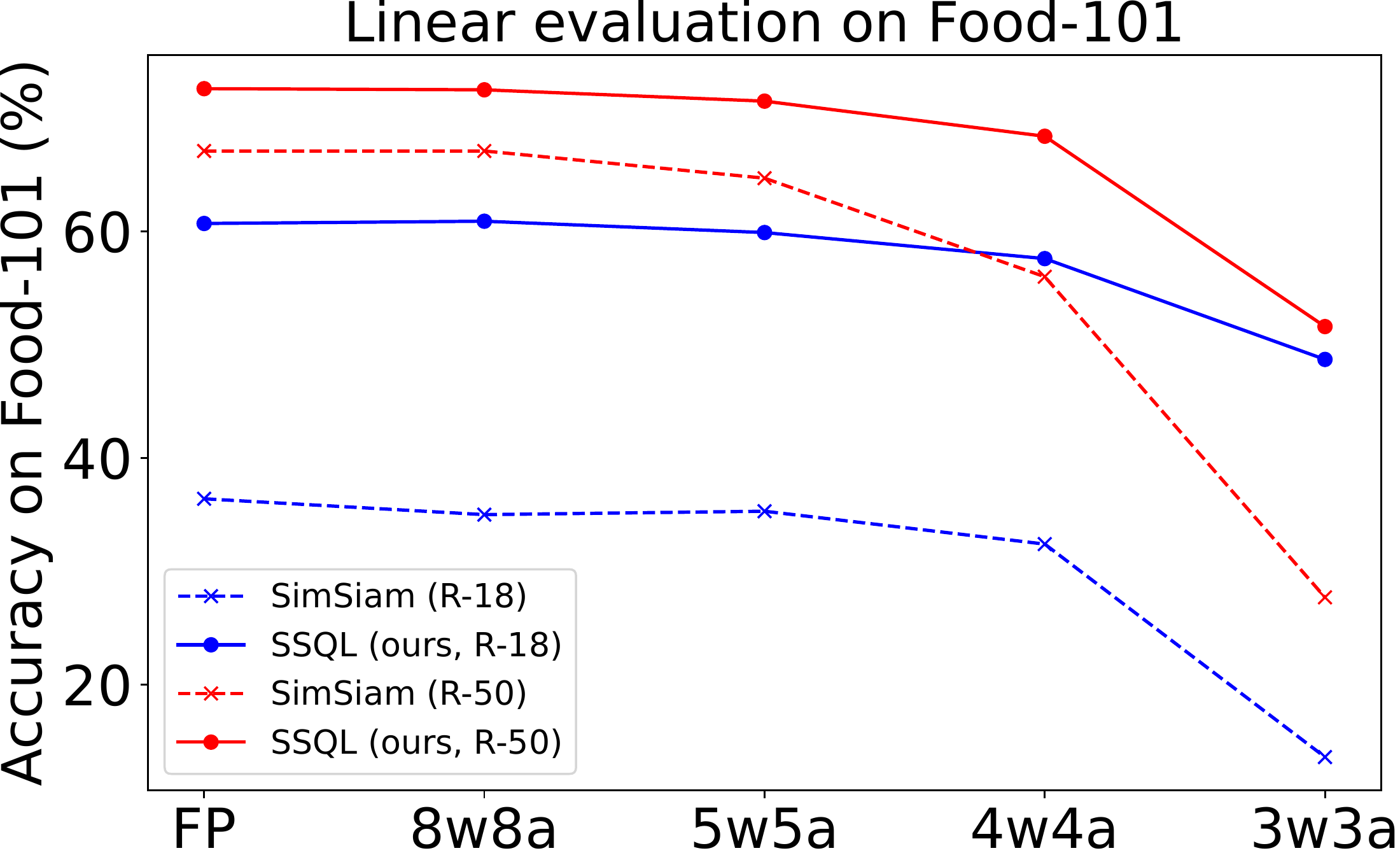}
	}
    \subfloat[Pets+Lin]{
        \label{fig:appendix-pets-linear}
        \includegraphics[width=0.32\linewidth]{dtd_linear_eval.pdf}
    }
    \\
    \subfloat[CIFAR-100+FT]{
        \label{fig:appendix-cifar100-finetune}
        \includegraphics[width=0.32\linewidth]{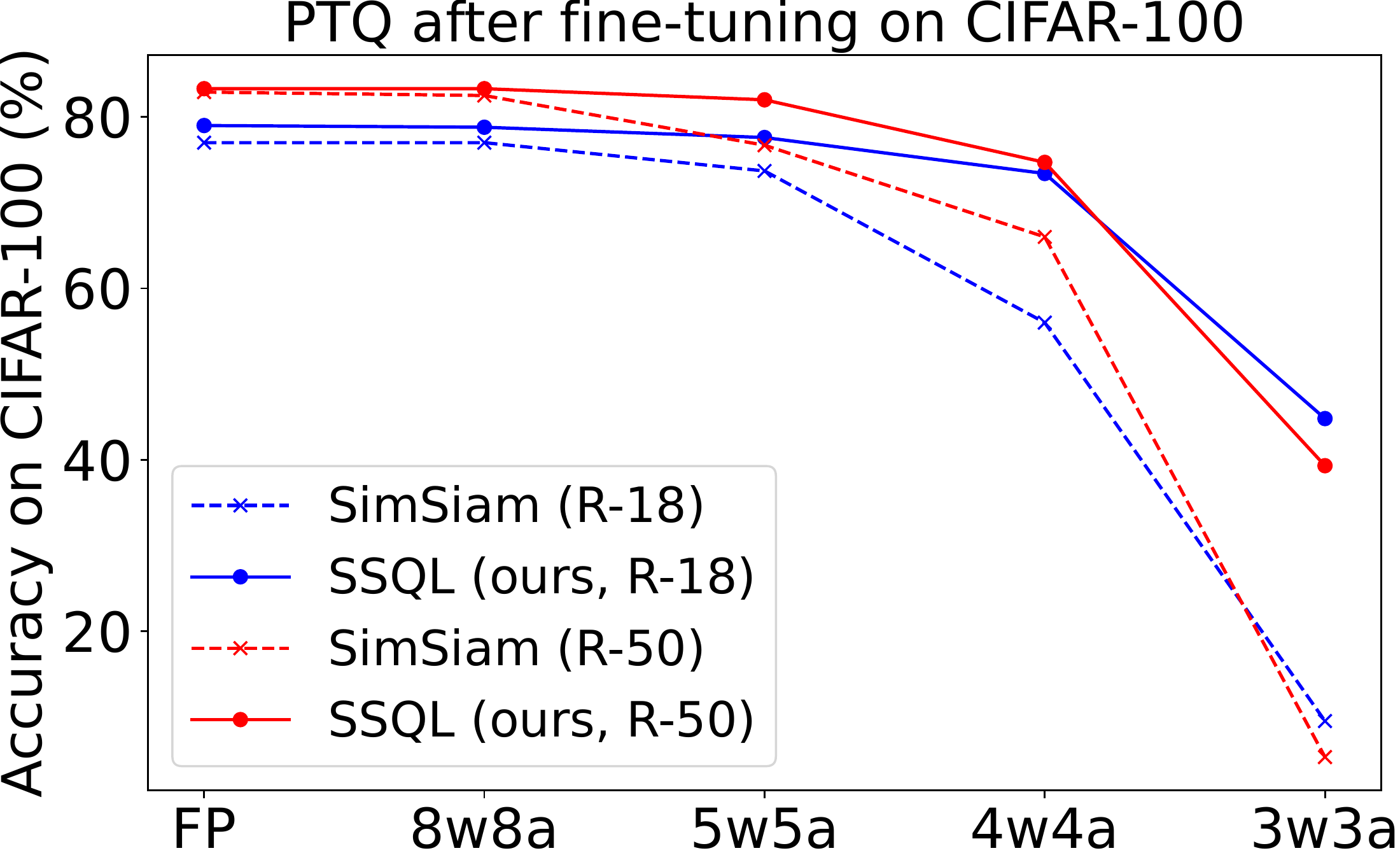}
	}
    \subfloat[Food-101+FT]{
        \label{fig:appendix-food-finetune}
        \includegraphics[width=0.32\linewidth]{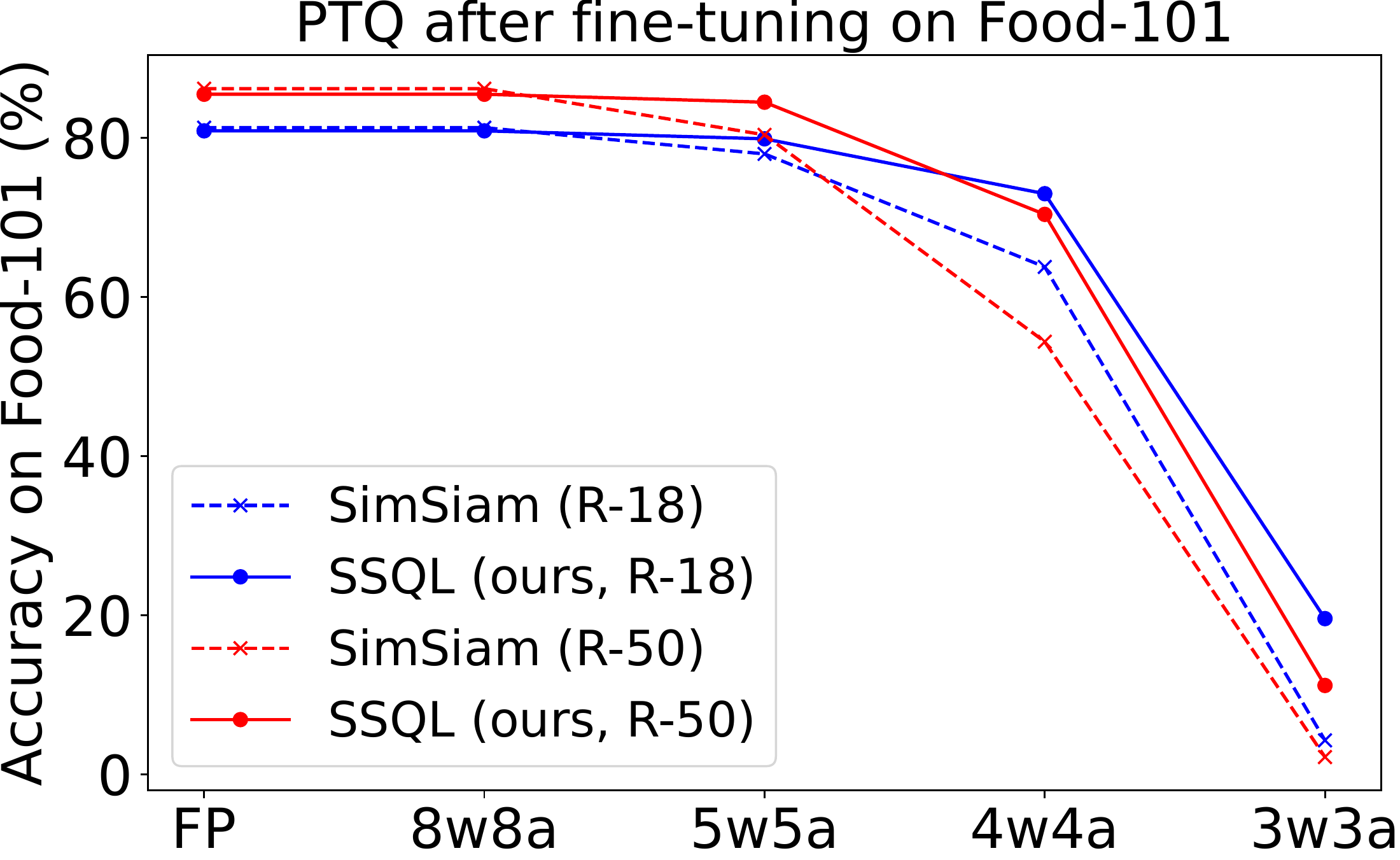}
	}
    \subfloat[Pets+FT]{
        \label{fig:appendix-pets-finetune}
        \includegraphics[width=0.32\linewidth]{dtd_finetune.pdf}
    }
    \caption{Transfer classification results. The first and second row shows the results under linear evaluation and fine-tuning (`FT'), respectively. Best viewed in color.}
    \label{fig:appendix-transfer-classification}
\end{figure*}

\subsection{Transfer results}\label{sec:app:transfer}
\noindent\textbf{Classification benchmarks.} We present the ImageNet transfer results on classification benchmarks under ResNet-18 in Table~\ref{tab:transfer_classification_r18}. We can see that our SSQL not only greatly improves the performance when quantized to low bit-widths, but also improves the performance of full precision models in some cases.

\noindent\textbf{Combining with LSQ on COCO.} We initialize LSQ with COCO fine-tuned models. Notice that we only quantize the backbone here (without quantizing ROI heads). As shown in Fig.~\ref{fig:LSQ-coco}, we can observe that our SSQL provides a better starting point for low bit QAT training on COCO. Take 2w4a ($\text{AP}_{\text{bb}}$) as an example, SSQL achieves 6.3 points higher than SimSiam (25.4 v.s. 19.1) after the first 1k iteration, while the initial accuracy of the FP model is about the same (38.2 v.s. 38.4). Consequently, our SSQL achieves higher final accuracy (36.4 v.s. 35.8) and it shows that our pretrained model can serve as a better initialization when combined with QAT methods to boost performance.

\subsection{Applications in other SSL methods}\label{sec:app:ssql_mocov2}
In this subsection, we demonstrate that our method SSQL can also work on other SSL frameworks. We experiment on MoCov2 and BYOL on CIFAR-10 under R-18 in Table~\ref{tab:app-cifar10-mocov2}. Our SSQL has consistent improvements, too.
\begin{table}
    \caption{Linear evaluation results on CIFAR-10.}
    \label{tab:app-cifar10-mocov2}
    \centering
    \begin{tabular}{l|c|c|c|c|c|c|c}
    \hline
     Backbone&Method&FP&6w6a&5w5a&4w4a&3w3a&2w4a \\
     \hline
     \multirow{4}{*}{ResNet-18}&BYOL& 89.3 & 89.4 & 89.3 & 88.0 & 75.1 & 63.3  \\
     &BYOL+SSQL&\textbf{90.8} & \textbf{90.7} & \textbf{90.6} & \textbf{89.8} & \textbf{85.0} & \textbf{85.7}\\
     &MoCov2& 88.9 & 88.4 & 88.2 & 86.8& 72.2 & 50.7 \\
     &MoCov2+SSQL& \textbf{89.6} & \textbf{89.6} & \textbf{89.5}& \textbf{88.5} &\textbf{83.4} & \textbf{85.2} \\
     \hline
    \end{tabular}
\end{table}

\subsection{Applications in vision transformers}\label{sec:app:ssql_vit}
In this subsection, we investigate whether the SSQL can be applied on Transformer-like backbones to achieve effectiveness. We supplement the results on CIFAR-10 using ViT-Small by adapting SSQL to MoCov3~\cite{mocov3:chen:ICCV21} (we use official code and conduct linear evaluation) in Table~\ref{tab:app-mocov3}. Our SSQL does have potentials on Transformer-like backbones.

\begin{table}
    \caption{Linear evaluation results on CIFAR-10 under ViT-Small.}
    \label{tab:app-mocov3}
    \centering
    \begin{tabular}{l|c|c|c|c|c|c}
    \hline
     Backbone & Method &FP&8w8a&6w6a&5w5a&4w4a\\
     \hline
     \multirow{2}{*}{ViT-Small} & MoCov3&  88.0 &87.6 & 87.2 &82.2&82.0  \\
     & MoCov3+SSQL&  \textbf{88.6} & \textbf{88.6} & \textbf{88.3} & \textbf{88.2} &\textbf{86.9}\\
     \hline
    \end{tabular}
\end{table}

\clearpage
\section{More analysis of the synergy}
In this section, we give more analysis as a supplement to Sec.~\ref{sec:theory} in the paper. We give empirical support for the weakly correlated assumption in Sec.~\ref{sec:app:assumption} and analyze the synergy from another perspective in Sec.~\ref{sec:app:another}.

\subsection{Support for the weakly correlated assumption} \label{sec:app:assumption}

\begin{figure}[htbp]
	\centering
	\includegraphics[width=0.47\columnwidth]{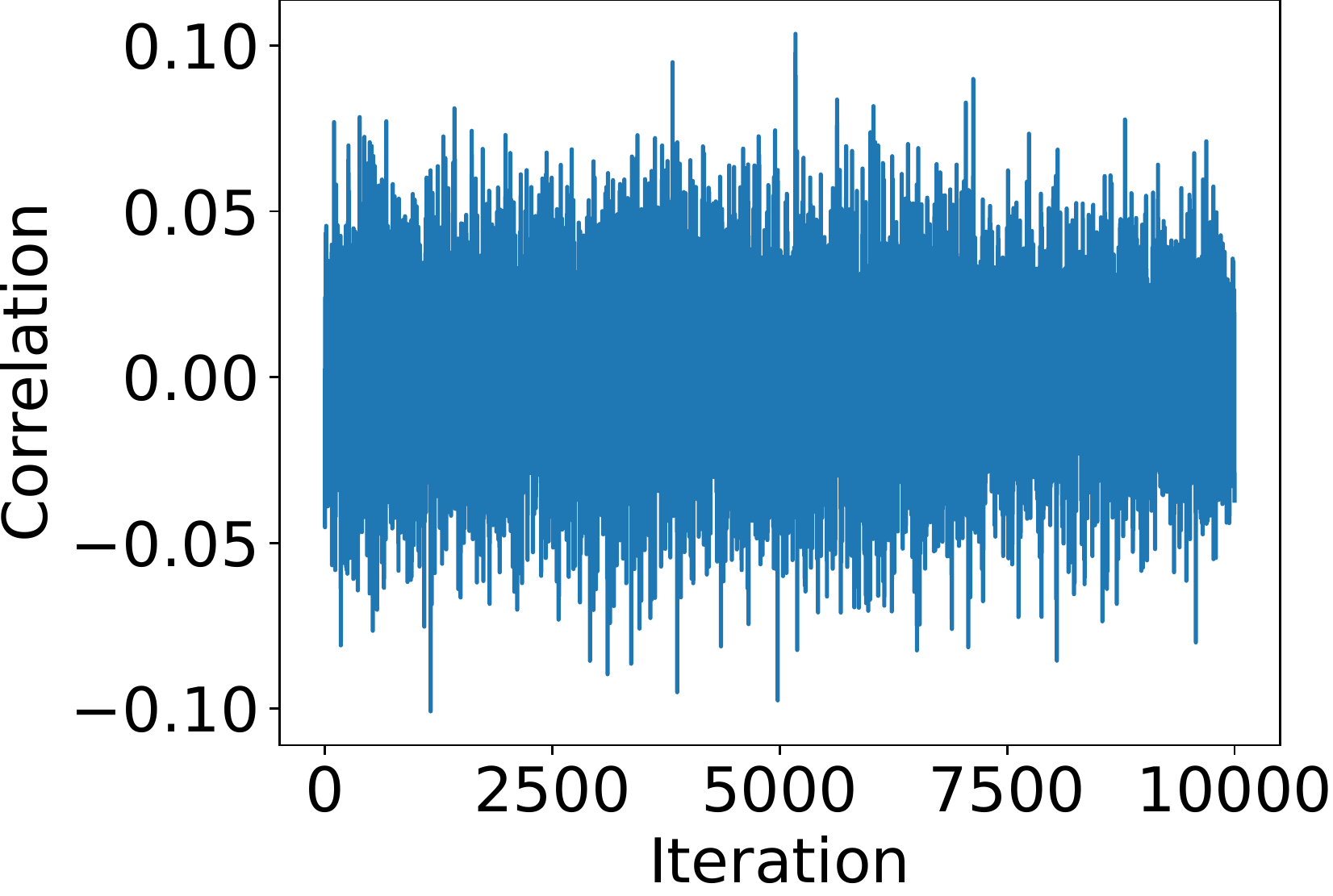}
	\quad
	\includegraphics[width=0.43\columnwidth]{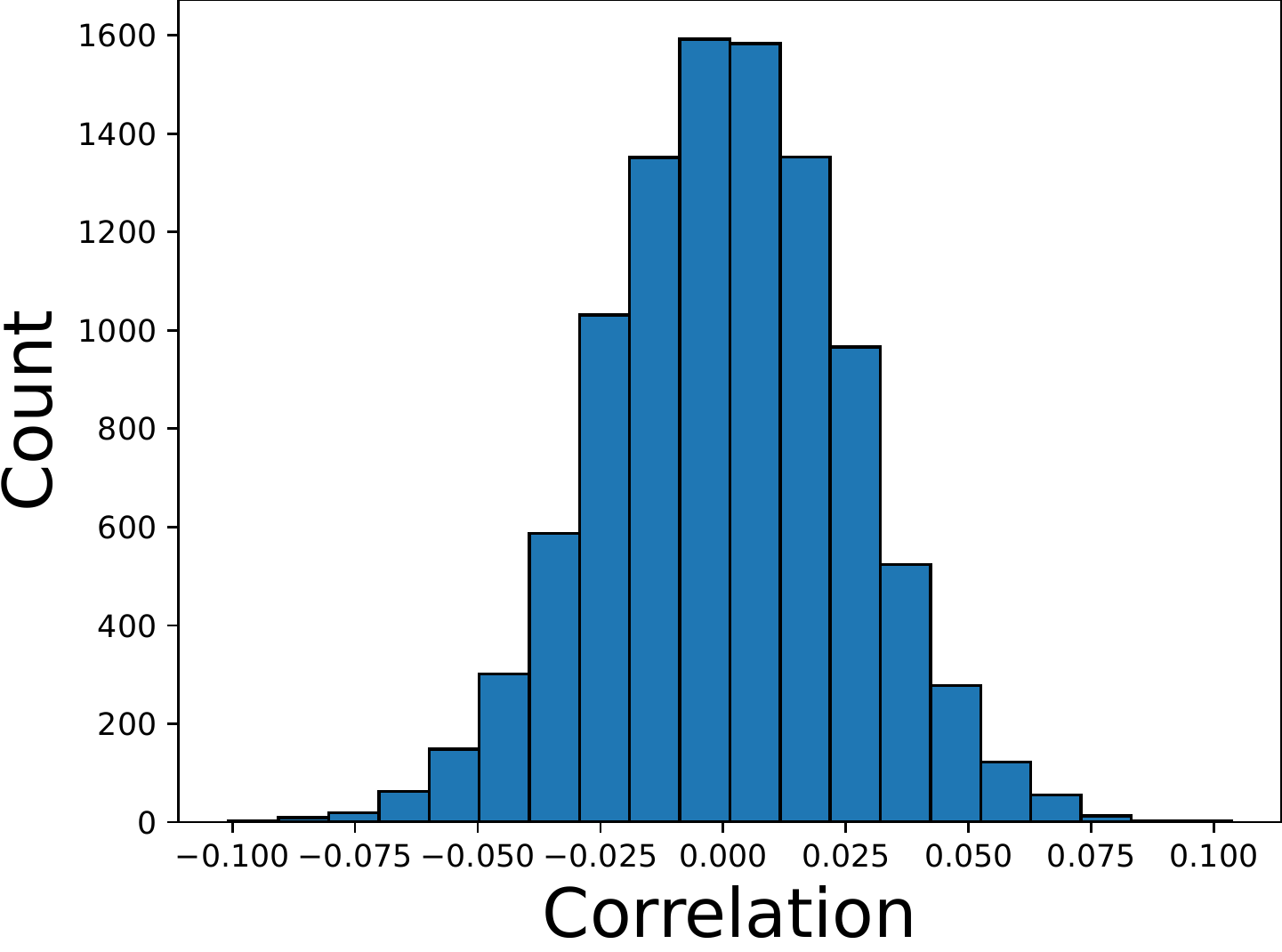}
    \caption{The correlation between the quantization and contrastive errors during training on CIFAR-10 using ResNet-18. Left: Curve of the correlation. Right: Histogram distribution of the correlation.}
	\label{fig:app:correlation}
\end{figure}

We plot the curve and histogram of correlation between the quantization and contrastive errors during training (10k iterations$\approx$100 epochs) in Fig.~\ref{fig:app:correlation}. Notice that the value range is $[-1,1]$ and in most cases the correlation is around 0 (i.e., uncorrelated), and it does not exceed $\pm$0.1. Experimental results verify that our assumption is a reasonable one.

\subsection{Analysis of the synergy from another perspective}
\label{sec:app:another}

For simplicity, we assume $f$ is a two-layer perceptron: 

\begin{equation}
    \boldsymbol{z}_1=\boldsymbol{w}_2\sigma(\boldsymbol{w}_1 \cdot \boldsymbol{x}_1) \,,
\end{equation}
where $\boldsymbol{w}_1$ and $\boldsymbol{w}_2$ are the corresponding weights and $\sigma(\cdot)$ is the activation function.

We consider only the first term in each loss function (i.e., the similarity between $\boldsymbol{p}_1$ and $\boldsymbol{z}_2$) without loss of generality. Suppose we quantize $\boldsymbol{w}_2$ and $\boldsymbol{w}_2^q = \boldsymbol{w}_2 + \Delta w$ and the analysis is the same for other weights or activations.

\begin{align*}
    L_{SSQL} =& -p_1^q \cdot z_2= -h(z_1^q) \cdot z_2 \\
        =& -h(z_1+\Delta z) \cdot z_2 \\
        \approx & - \Big(h(z_1)+h'(z_1)\Delta z\Big) \cdot z_2\\
        = & -p_1\cdot z_2 - h'(z_1) \Delta z \cdot z_2\,,\\
\end{align*}
where $\Delta z = \Delta w\sigma(w_1\cdot x_1)$. By introducing quantization noise, we can see that its effect can be thought of as adding a random perturbation to the points before evaluating their similarity as usual. This provides an explanation on why our method could lead to better results.

We now investigate the backward pass for $L_{SSQL}$:
\begin{equation}
\label{eq:SSQL_z1}
\begin{aligned}
    \frac{\partial L_{SSQL}}{\partial z_1} =&  \frac{\partial L}{\partial p_1^q} \cdot \frac{\partial p_1^q}{\partial z_1}\\
    =& -z_2 \cdot \frac{\partial h(z_1+\Delta z)}{\partial z_1} \\
    \approx & -z_2 (h'(z_1)+h''(z_1) \Delta z)
\end{aligned}
\end{equation}

\begin{equation}
\frac{\partial L_{SSQL}}{\partial w_2} =  \frac{\partial L}{\partial z_1} \cdot \frac{\partial z_1}{\partial z^q_1} \cdot \frac{\partial z_1^q}{\partial w_2} = \frac{\partial L}{\partial z_1} \cdot \sigma(w_1\cdot x_1)
\end{equation}

\begin{equation}
    \frac{\partial L_{SSQL}}{\partial w_1} =  \frac{\partial L}{\partial z_1} \cdot \frac{\partial z_1}{\partial z^q_1} \cdot \frac{\partial z^q_1}{\partial w_1}\\
    =  \frac{\partial L}{\partial z_1} \cdot (w_2+\Delta w)\sigma'(w_1\cdot x_1) \cdot x_1
\end{equation}

The backward pass for $L_{SimSiam}$ is obvious from the above derivations:

\begin{equation}
\label{eq:simsiam_z1}
    \frac{\partial L_{SimSiam}}{\partial z_1} = -z_2 h'(z_1)  
\end{equation}

\begin{equation}
    \frac{\partial L_{SimSiam}}{\partial w_1} = -\frac{\partial L_{SimSiam}}{\partial z_1} \cdot w_2 \sigma'(w_1\cdot x_1) \cdot x_1  
\end{equation}

When comparing Equation~\eqref{eq:SSQL_z1} with Equation~\eqref{eq:simsiam_z1}, we can find an extra term $-z_2 h''(z_1)\Delta z$ in the gradients and we think this second-order term can help model learn better.

\end{document}